\documentclass[lettersize,journal]{IEEEtran}

\def\BibTeX{{\rm B\kern-.05em{\sc i\kern-.025em b}\kern-.08em
    T\kern-.1667em\lower.7ex\hbox{E}\kern-.125emX}}

\usepackage{cite}
\usepackage{amsmath,amssymb,amsfonts}
\usepackage{graphicx}
\usepackage{tikz}
\usetikzlibrary{arrows.meta,positioning,calc}
\usepackage{textcomp}
\usepackage{xcolor}
\usepackage{hyperref}
\usepackage{booktabs}
\usepackage{array}
\usepackage[caption=false,font=normalsize,labelfont=sf,textfont=sf]{subfig}
\usepackage{stfloats}
\usepackage{balance}
\usepackage{placeins}
\usepackage[font=small, labelfont=bf]{caption}
\usepackage{float}
\usepackage{bm}
\usepackage{xspace}
\usepackage{soul}
\usepackage{circledtext}
\usepackage[ruled,vlined,linesnumbered]{algorithm2e}
\usepackage{multirow}
\usepackage{paralist}
\usepackage{tcolorbox}
\usepackage{siunitx}
\usepackage{framed}

\newcommand{\method}{\texttt{U-Balance}\xspace}
\newcommand{\ulnr}{\texttt{uLNR}\xspace}
\DeclareSIUnit{\pp}{\textit{pp}}
\newcommand\rev[1]{\textcolor{black}{#1}}
\begin{document}

\title{ Uncertainty-Guided Label Rebalancing for CPS Safety Monitoring}
\author{John Ayotunde,~\IEEEmembership{}
        Qinghua Xu,~\IEEEmembership{}
        Guancheng Wang,~\IEEEmembership{}
        and~Lionel~C.~Briand~\IEEEmembership{}
\thanks{J. Ayotunde, Q. Xu, and G. Wang are with Lero Research Ireland Centre for Software Research, University of Limerick, Castletroy, Limerick, Ireland. E-mail: \{ayotunde.johnoluwatobiloba, qinghua.xu, guancheng.wang\}@ul.ie}
\thanks{L. C. Briand is with the University of Ottawa, Canada, and Lero Research Ireland Centre for Software Research, University of Limerick, Ireland. E-mail: Lionel.Briand@lero.ie}}


\maketitle

\begin{abstract}
Safety monitoring is essential for Cyber-Physical Systems (CPSs) such as Unmanned Aerial Vehicles (UAVs). However, unsafe events are rare in real-world CPS operations, creating an extreme class imbalance that degrades data-driven safety predictors. Meanwhile, behavioral uncertainty in CPS operations, defined as the degree of doubt or uncertainty in CPS decisions (e.g., erratic control signals or rapid heading changes), is often correlated with safety outcomes:  Uncertain behaviors are more likely to occur during unsafe operation. However, safety monitoring has underexplored this valuable uncertainty information.

To that end, we propose \method, a supervised approach that leverages behavioral uncertainty to rebalance imbalanced datasets prior to training a safety predictor. \method first trains a GatedMLP-based uncertainty predictor that summarizes each telemetry window into distributional kinematic features and outputs an uncertainty score. It then applies an uncertainty-guided label rebalancing (\ulnr) mechanism that probabilistically relabels \textit{safe}-labeled windows with unusually high uncertainty as \textit{unsafe}, thereby enriching the minority class without synthesizing new data. Finally, we train a safety predictor on the rebalanced dataset for safety monitoring.

We evaluate \method on a UAV benchmark and on ForEnt, a quadruped-robot entrapment dataset. \rev{On UAV, \method achieves F1/AUROC/AUPRC of $0.801/0.972/0.776$, outperforming the strongest safety-prediction baselines by 14.3/2.2/6.1 percentage points (\textit{pp}), respectively. While the strongest baseline varies across metrics (MLP for F1 and AUPRC, and CNN for AUROC), \method consistently achieves the highest scores across all three metrics. Similarly, on ForEnt, \method achieves F1/AUROC/AUPRC of $0.522/0.806/0.323$, outperforming the strongest baselines by 4.4/9.2/1.4 \textit{pp}. With the same Bi-LSTM safety predictor, the gains over the strongest alternative rebalancing methods are 17.0/4.1/8.2 percentage points on UAV and 2.3/6.5/1.9 on ForEnt.} \ulnr outperforms Early Fusion and Late Fusion on both datasets. To our knowledge, this work is the first to use behavioral uncertainty for dataset rebalancing in CPS safety monitoring.

\end{abstract}


\maketitle

\section{Introduction}
Cyber-Physical Systems (CPSs), such as Unmanned Aerial Vehicles (UAVs), are increasingly deployed across diverse application scenarios~\cite{superialist}, including crop monitoring~\cite{crop} and disaster rescue~\cite{rescue}. These capabilities have made CPSs more complex, which in turn broadens their safety attack surface and exacerbates safety threats~\cite{duo2022survey,d2014guest,di2023automated}. Indeed, UAV crashes and operational failures have caused property damage and endangered people~\cite{incident1, incident2, incident3,incident4}, underscoring the importance of CPS safety. 

Safety monitoring is a vital strategy for detecting and preventing unsafe behaviors during CPS operation~\cite{jiang2018data}. By continuously assessing runtime behavior, monitors can detect deviations from expected norms and enable timely interventions such as human takeover~\cite{takeover} or fail-safe modes~\cite{fallback1,fallback2}. Existing monitors are broadly rule-based, explicitly encoding safety constraints~\cite{rule1,rule2,rule3}, or data-driven, using machine learning (ML) to infer safety-related problems from real-time sensor/actuator telemetry~\cite{stocco2020misbehaviour,xu2021digital,xu2023digital,michelmore2020uncertainty}.

Supervised safety predictors rely on labeled datasets but are sensitive to label imbalance. In practice, safety-monitoring data are often extremely imbalanced, with safe samples vastly outnumbering unsafe ones; for instance, the safe-to-unsafe ratio is approximately $76\!:\!1$ in the UAV training data used in this study~\cite{superialist}. Models trained under such imbalance tend to over-predict the majority class and fail to learn representative patterns from rare yet critical unsafe behavior. To mitigate this issue, common rebalancing techniques include synthetic over-sampling (e.g., SMOTE)~\cite{chawla2002smote} and loss reweighting (e.g., class weighting)~\cite{bakirarar2023class,zhu2018class}. More recently, label noise rebalancing (LNR) has emerged as a promising alternative that rebalances labels without synthesizing any samples, even under extreme imbalance~\cite{lnr}. LNR stochastically flips selected majority-class labels to the minority class, guided by a classifier's predicted probabilities.

Despite its effectiveness, directly applying LNR to CPS time-series safety monitoring is non-trivial. Prior evaluations of LNR focus on tabular and image data, whereas CPS telemetry is sequential and temporally dependent: unsafe behavior often arises from evolving patterns over time rather than isolated feature values. Moreover, CPSs often operate in open contexts~\cite{opencontext1,opencontext2}, where environmental conditions and system dynamics drift. These challenges motivate a time-series-aware rebalancing strategy that selects informative majority-class samples based on behavioral characteristics, rather than relying solely on a safety classifier's predictions.

Meanwhile, recent work by Khatiri et al.~\cite{superialist} shows that \emph{behavioral uncertainty} in CPS operation—manifested as erratic control signals or rapid heading changes—is correlated with safety outcomes: uncertain behaviors occur more frequently during unsafe operation. This suggests that uncertainty can serve as a valuable signal for improving safety prediction. However, how to effectively leverage behavioral uncertainty to enhance safety monitoring remains underexplored.

To that end, we propose \textbf{\method}, a safety-monitoring approach that leverages behavioral uncertainty to adapt LNR to time-series CPS data. \method first trains an uncertainty predictor that summarizes each telemetry window into distributional kinematic features and outputs an uncertainty score that captures behavioral uncertainty (e.g., rapid changes in UAV heading angles). It then applies an uncertainty-guided LNR mechanism (\ulnr) that probabilistically relabels \textit{safe}-labeled windows with unusually high predicted uncertainty (relative to other \textit{safe} windows) as \textit{unsafe}, enriching the minority class without synthesizing new data. Finally, \method trains a safety predictor on the rebalanced dataset for runtime safety monitoring.

\rev{We evaluate \method on the UAV benchmark constructed by Khatiri et al.~\cite{superialist} and the ForEnt quadruped-robot dataset~\cite{forent}. Predicted behavioral uncertainty is moderately correlated with safety on both datasets. Using the same predicted uncertainty scores, \ulnr outperforms Early Fusion and Late Fusion. \method achieves F1/AUROC/AUPRC of $0.801/0.972/0.776$ on UAV and $0.522/0.806/0.323$ on ForEnt. In end-to-end comparisons against safety-prediction baselines, \method improves F1/AUROC/AUPRC over the strongest baseline for each metric by 14.3/2.2/6.1 percentage points (\textit{pp}) on UAV (MLP for F1 and AUPRC, and CNN for AUROC) and by 4.4/9.2/1.4~\textit{pp} on ForEnt (CNN, Transformer, and Bi-GRU, respectively). The best baseline varies across metrics, so the reported gains are conservative: comparing with any single baseline would yield larger gains on some metrics. These results show that \method delivers the best overall safety-prediction performance, rather than performing well on only one metric. When all rebalancing methods use the same Bi-LSTM safety predictor, the corresponding gains are 17.0/4.1/8.2~\textit{pp} on UAV and 2.3/6.5/1.9~\textit{pp} on ForEnt.}

In summary, our contributions are threefold:
\begin{compactenum}
\item We propose \method, introducing \ulnr for CPS safety monitoring. \ulnr rebalances imbalanced time-series datasets by using predicted behavioral uncertainty to identify informative majority-class samples for relabeling. To the best of our knowledge, \method is the first to exploit behavioral uncertainty for dataset rebalancing, improving data-driven safety monitoring and offering an alternative to standard uncertainty-fusion strategies.

\item We propose an uncertainty predictor for time-series telemetry that summarizes each window using distributional kinematic features and estimates behavioral uncertainty with a GatedMLP architecture. This design enables \method to adapt LNR to time-series data effectively.

\item \rev{We evaluate \method on UAV and ForEnt, selected because they provide safety annotations, record CPS operation, and capture behavioral uncertainty in telemetry. \method achieves the highest F1, AUROC, and AUPRC among safety-prediction baselines on both datasets. \ulnr also achieves the highest F1, AUROC, and AUPRC among alternative rebalancing methods on both datasets.}
\end{compactenum}

\section{Background}
\label{sec:background}

\subsection{CPS Operation, Safety, and Uncertainty}
\noindent\textbf{CPS Operation.} \rev{A CPS combines software with a physical system through sensors and actuators. Sensors report the system's state, software computes control actions, and actuators change its physical behavior. UAVs and quadruped ground robots are two examples of this feedback process.}

\begin{figure}
    \centering
    \includegraphics[width=0.93\linewidth]{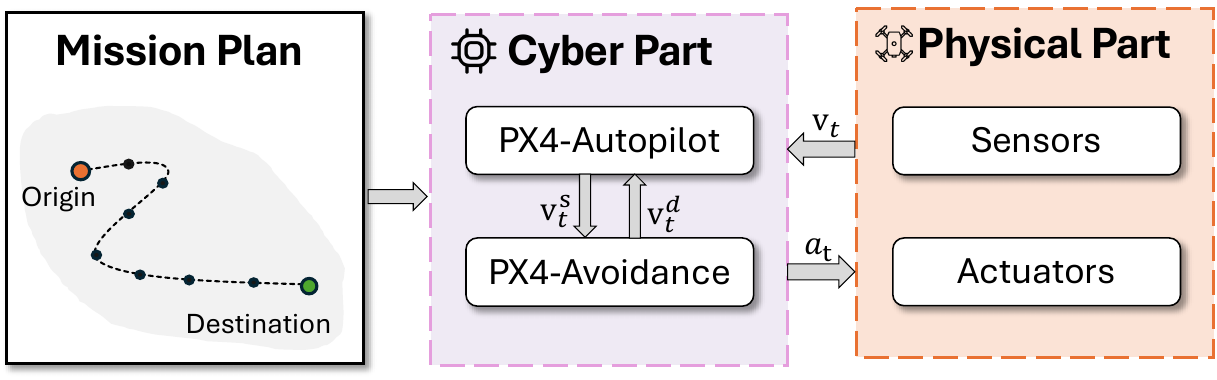}
    \caption{Example of a UAV operation.}
    \label{fig:uav-operation}
\end{figure}

\rev{A UAV follows a mission plan while responding to its surroundings~\cite{superialist}. In Figure~\ref{fig:uav-operation}, onboard sensors provide the current position and heading $v_t=(x_t,y_t,z_t,r_t)$. The autopilot computes the desired state $v_t^d$, and the obstacle-avoidance module adjusts it to produce $v_t^s$. The autopilot then translates this state into commands $a_t$ that control the UAV's motion.}

\begin{figure}[t]
    \centering
    \includegraphics[width=\linewidth]{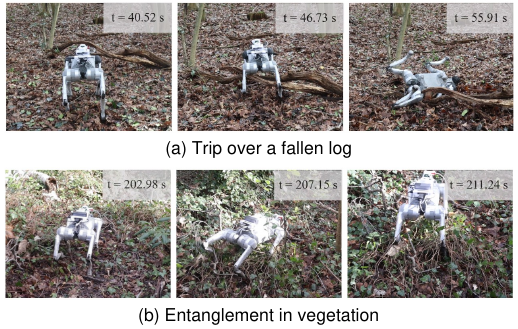}
    \caption{\rev{Examples of Go2 robot operation on forest terrain. (a) Approaching and tripping over a fallen log. (b) Moving through vegetation and becoming entangled. Go2 robot photos provided by Kirdwichai and Tarapore~\cite{forent}.}}
    \label{fig:robot-operation}
\end{figure}

\rev{A quadruped robot follows movement commands through coordinated leg actuation. In ForEnt, a teleoperator guides a Unitree Go2 through forest terrain, while its locomotion controller regulates the leg movements~\cite{forent}. Body acceleration, angular velocity, and orientation record how the robot responds to the terrain. Figure~\ref{fig:robot-operation} shows examples of these physical interactions.}

\noindent\textbf{Safety.} \rev{Safe operation requires a CPS to satisfy the physical constraints of its task. For UAVs, these can include obstacle clearance, altitude and speed limits, and reliable control. We focus on collision-related safety: flying too close to an obstacle creates a collision risk~\cite {superialist,surrealist}. For ground robots, maintaining mobility and balance is essential. Vegetation can entangle the legs, rigid obstacles can cause trips, and soft ground can cause the feet to sink~\cite{forent}. These interactions can lead to \emph{entrapment}, where the robot is stuck despite receiving a movement command, or to a loss of balance and a fall. The two case studies therefore examine different physical safety concerns within the same CPS monitoring problem.}

\noindent\textbf{Behavioral Uncertainty.} \rev{Sensor noise, changing environments, and unexpected obstacles can disrupt consistent CPS behavior. These sources include aleatoric uncertainty from inherent randomness and epistemic uncertainty from incomplete knowledge or models~\cite{abdar2021review}. Following Khatiri et al.~\cite{superialist}, we use \emph{behavioral uncertainty} to describe observable irregularities in the system's motion or control. Examples include erratic heading corrections in UAVs and irregular body acceleration or rotation in ground robots. We characterize this observed behavior from telemetry without separating its aleatoric and epistemic sources.}

\rev{We emphasize two distinctions. First, behavioral uncertainty describes a CPS's observed behavior, whereas methods such as MC Dropout~\cite{gal2016dropout} and Deep Ensembles~\cite{lakshminarayanan2017simple} estimate uncertainty in a predictive model's output. Second, behavioral uncertainty differs from unsafety.}

\begin{figure}
    \centering
    \includegraphics[width=1\linewidth]{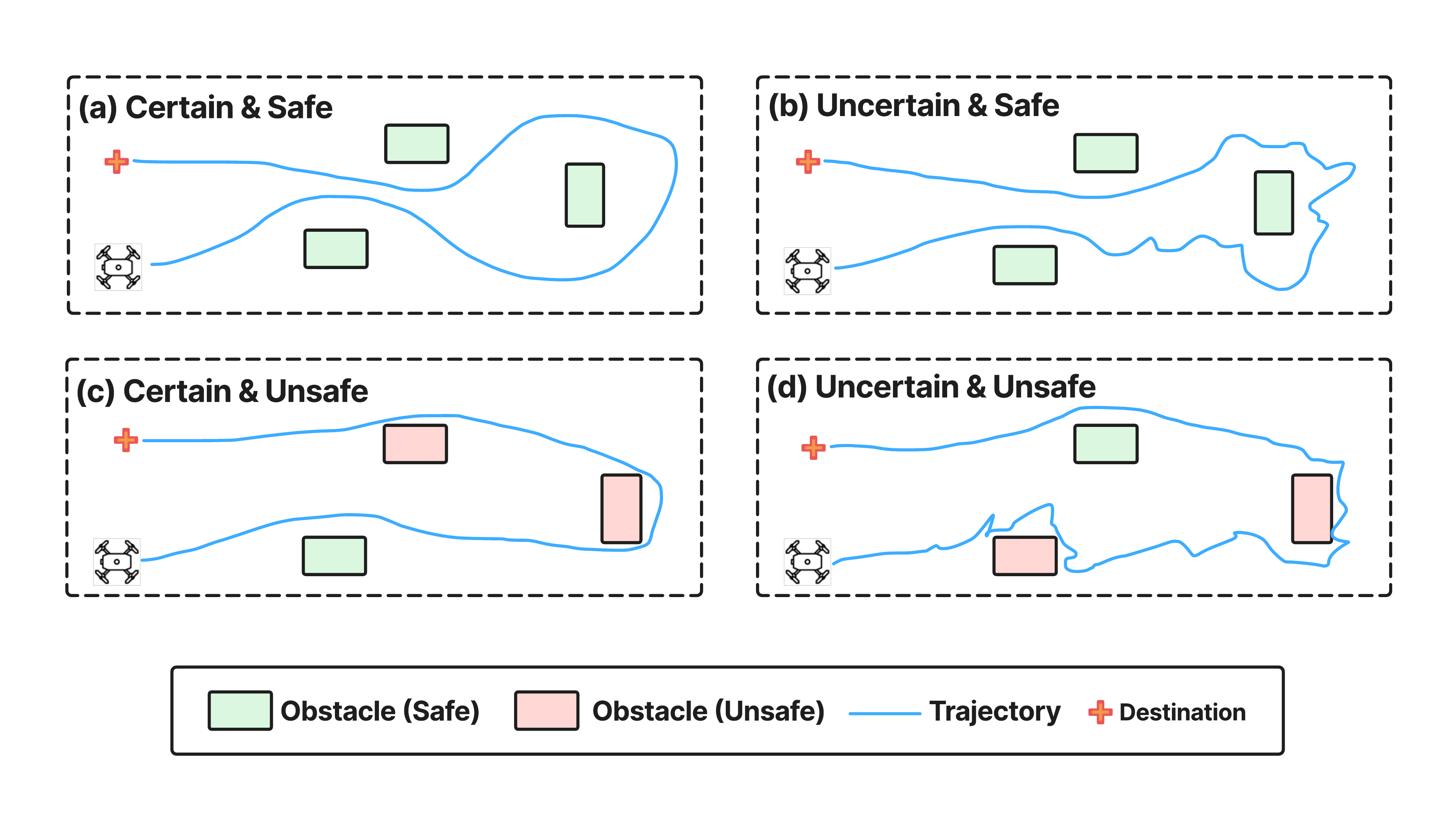}
    \caption{Examples of uncertainty and safety in UAV operations.}
    \label{fig:flight-types}
\end{figure}

\rev{To illustrate the second distinction, Figure~\ref{fig:flight-types} shows four UAV examples: (a) smooth flight with sufficient clearance, (b) erratic heading corrections with sufficient clearance, (c) smooth flight with insufficient clearance, and (d) erratic heading corrections with insufficient clearance. Both smooth and erratic flight can therefore be safe or unsafe, depending on obstacle clearance. The same distinction applies to ground robots: a robot can make irregular corrective movements while still progressing through terrain, or remain stuck with little body motion.}

\subsection{Safety Monitoring in CPSs}
\rev{Safety monitoring predicts unsafe CPS states from runtime telemetry so that timely interventions can be triggered. We formulate this as binary classification of telemetry windows. Let $\omega_t\in\mathbb{R}^{T\times C}$ denote a window ending at time $t$, where $T$ is the number of time steps and $C$ is the number of telemetry channels. Each window has a safety label $l_t^s\in\{0,1\}$, with $0$ denoting safe and $1$ denoting unsafe. The safety criteria and labeling rules for UAV and ForEnt are described in Section~\ref{subsec:dataset}.}

The training set is
\begin{equation}
\label{eq:safety-dataset}
\mathcal{D} = \{(\omega_t,l_t^s)\}_{t=1}^{N},
\end{equation}
and the objective is to learn a safety predictor
\begin{equation}
\label{eq:safety-predictor-objective}
f: \mathbb{R}^{T \times C} \rightarrow [0,1], \quad f(\omega_t) \approx P(l_t^s=1 \mid \omega_t),
\end{equation}
which estimates the probability that a window is unsafe. \rev{A key challenge is class imbalance: unsafe events are rare, so training data typically contain many more safe than unsafe windows. A predictor can consequently achieve high overall accuracy while failing to detect critical unsafe states.}

\section{Approach}
\label{sec:approach}

This section presents \method,  a supervised learning approach for safety monitoring in CPSs. The objective of \method is to train an effective safety predictor on a highly imbalanced dataset. \method is trained with both safety and uncertainty labels. For UAV, safety labels describe obstacle-related operation, while uncertainty labels come from expert annotations of behavioral consistency in flight trajectories~\cite{superialist}. ForEnt safety labels come from annotated entrapment intervals, and uncertainty targets are obtained through automated annotation based on horizontal-acceleration energy. Section~\ref{subsec:dataset} describes both datasets and their annotations.

\method consists of three components that operate collaboratively: \textcircled{1} \textit{Uncertainty Predictor}, \textcircled{2} \textit{\ulnr}, and \textcircled{3} \textit{Safety Predictor}. The training set is $\mathcal{D}=\{(\omega_t,l_t^s)\}_{t=1}^{N}$, where $N$ is the number of training windows. Uncertainty targets $l_t^u$ are available for an annotated subset of UAV training windows and are obtained through automated annotation based on heuristic rules for ForEnt. The uncertainty predictor is trained on windows with these targets and then assigns a score to every training window. \ulnr converts the scores into flip probabilities, and changes selected safe labels to unsafe, producing $\mathcal{D}_{bal}$. We train the safety predictor on $\mathcal{D}_{bal}$ for runtime monitoring.

In the rest of this section, we elaborate on each component: the Uncertainty Predictor (Section ~\ref{subsec:uncertainty-predictor}), \ulnr (Section ~\ref{subsec:lnr}), and the Safety Predictor (Section ~\ref{subsec:safety-predictor}). 

\subsection{Uncertainty Predictor}
\label{subsec:uncertainty-predictor}
\begin{figure}
    \centering
    \includegraphics[width=0.82\linewidth]{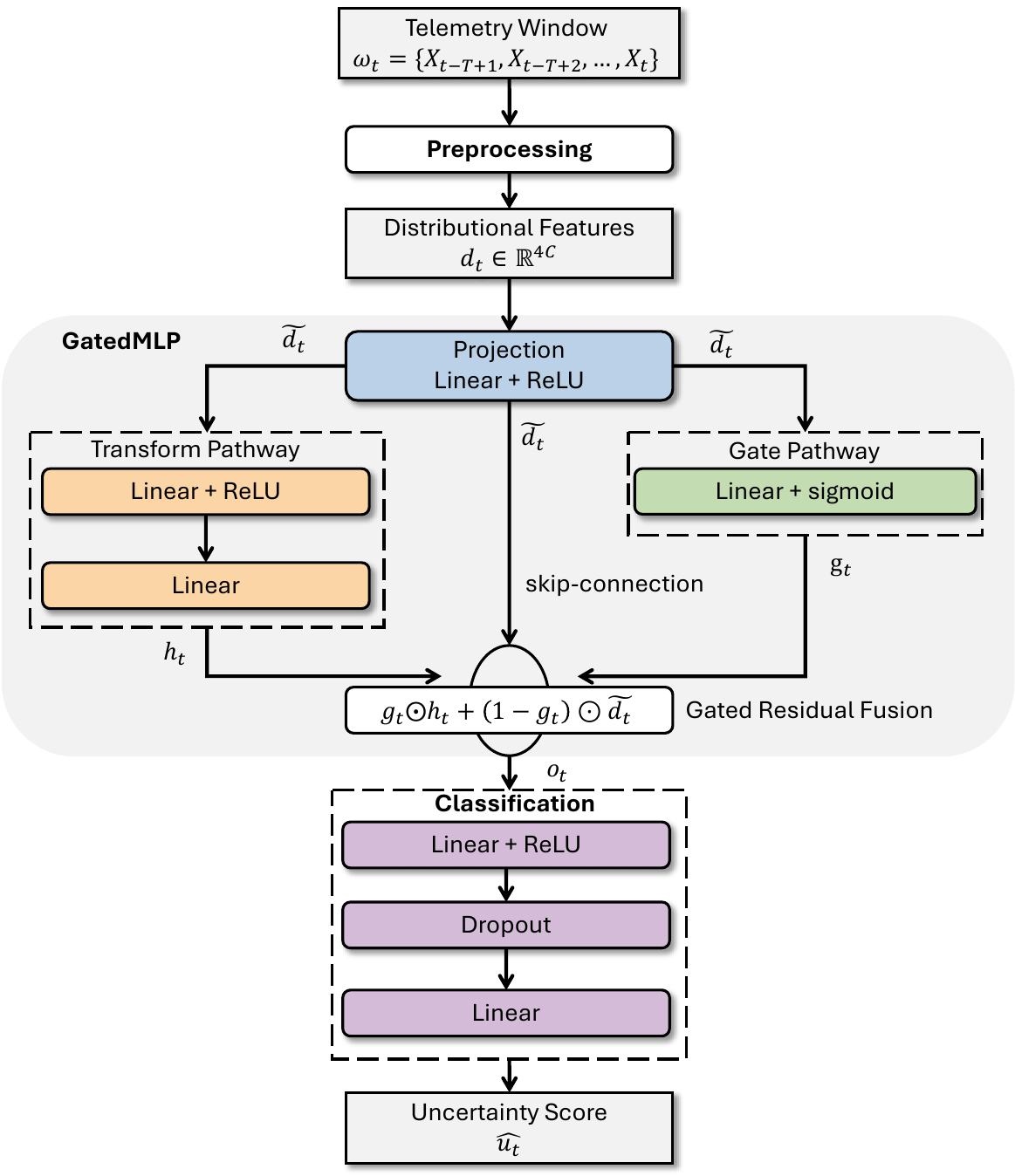}
    \caption{Overview of the Uncertainty Predictor (\textcircled{1}).}
    \label{fig:mu}
\end{figure}
The uncertainty predictor estimates behavioral uncertainty from CPS telemetry on both datasets. Figure~\ref{fig:mu} presents its architecture, consisting of three main steps: \textit{Preprocessing}, \textit{GatedMLP}, and \textit{Score prediction}.

\rev{Behavioral uncertainty can manifest as variability or extreme values in different telemetry channels. The relevant statistics can change across operating conditions. We therefore use distributional statistics to describe each window and GatedMLP to learn interactions among them. Its gate adjusts how much of each transformed feature is retained for a given window.}

\noindent\emph{\textbf{Preprocessing}.} Each window $\omega_t$ contains $T$ time steps and $C$ telemetry channels. We use the term \emph{channel} to denote one scalar variable measured at each time step (i.e., one dimension of the multivariate time series). UAV uses heading $r$ and position $(x,y,z)$. ForEnt uses three-axis acceleration, three-axis angular velocity, and roll, pitch, and yaw. For each channel $c$, we compute its mean, standard deviation, minimum, and maximum within the window:
\begin{equation}
\label{alg:preprocessing}
\begin{aligned}
d_t^{(c)} ={}& \bigl(\operatorname{mean}(\omega_t^{(c)}),
                    \operatorname{std}(\omega_t^{(c)}),\\
             &\quad \operatorname{min}(\omega_t^{(c)}),
                    \operatorname{max}(\omega_t^{(c)})\bigr),\\
d_t ={}& (d_t^{(1)},\ldots,d_t^{(C)})\in\mathbb{R}^{4C}.
\end{aligned}
\end{equation}
This produces 16 features for UAV and 36 for ForEnt. The statistics summarize motion variability and extreme values. For example, heading variation captures directional corrections on UAV, while acceleration and angular-rate variation describe body-motion variability on ForEnt.

\noindent\emph{\textbf{GatedMLP}.} The distributional feature vector $d_t$ is first passed to a linear layer $W_{\text{proj}}\in\mathbb{R}^{p\times4C}$ with bias $b_{\text{proj}}\in\mathbb{R}^p$, followed by a ReLU activation~\cite{relu} as in Equation~\ref{eq:gatedmlp_project}.
\begin{equation}
\label{eq:gatedmlp_project}
    \tilde{d}_t = \text{ReLU}(W_{\text{proj}} \, d_t + b_{\text{proj}}), \quad \tilde{d}_t \in \mathbb{R}^{p}
\end{equation}
Here, $p$ denotes the projection dimension. The projected representation $\tilde{d}_t$ then passes through a GatedMLP block, which uses a gating mechanism inspired by the Gated Recurrent Unit (GRU)~\cite{gru} to control information flow. Specifically, the block consists of two parallel pathways: a \emph{transform} pathway and a \emph{gate} pathway. The transform pathway uses an intermediate hidden layer to capture nonlinear interactions among features:
\begin{equation}
    h_t = W_2 \, \text{ReLU}(W_1 \, \tilde{d}_t + b_1) + b_2
\end{equation}
where $W_1 \in \mathbb{R}^{e \times p}$ further transforms the representation into an intermediate dimension $e > p$, and $W_2 \in \mathbb{R}^{p \times e}$ reduces it back. The gate pathway produces an element-wise gating vector:
\begin{equation}
    g_t = \sigma(W_g \, \tilde{d}_t + b_g), \quad g_t \in [0, 1]^{p}
\end{equation}
where $\sigma$ denotes the sigmoid function. The gate $g_t$ performs a learned element-wise interpolation between the transformed representation and the original projected input via a residual connection:
\begin{equation}
\label{eq:gated_residual}
    o_t = g_t \odot h_t + (1 - g_t) \odot \tilde{d}_t
\end{equation}

\noindent\emph{\textbf{Score prediction}.} The output $o_t$ passes through a prediction head with two linear layers, a ReLU activation, and dropout during training:
\begin{equation}
    \hat{u}_t = W_4 \,   \text{ReLU}(W_3 \, o_t + b_3)  + b_4
\end{equation}
Here, $\hat u_t\in\mathbb{R}$ is the predicted behavioral-uncertainty score used by \ulnr.

\subsection{Uncertainty-guided Label Rebalancing}
\label{subsec:lnr}
\begin{figure}
    \centering
    \includegraphics[width=1\linewidth]{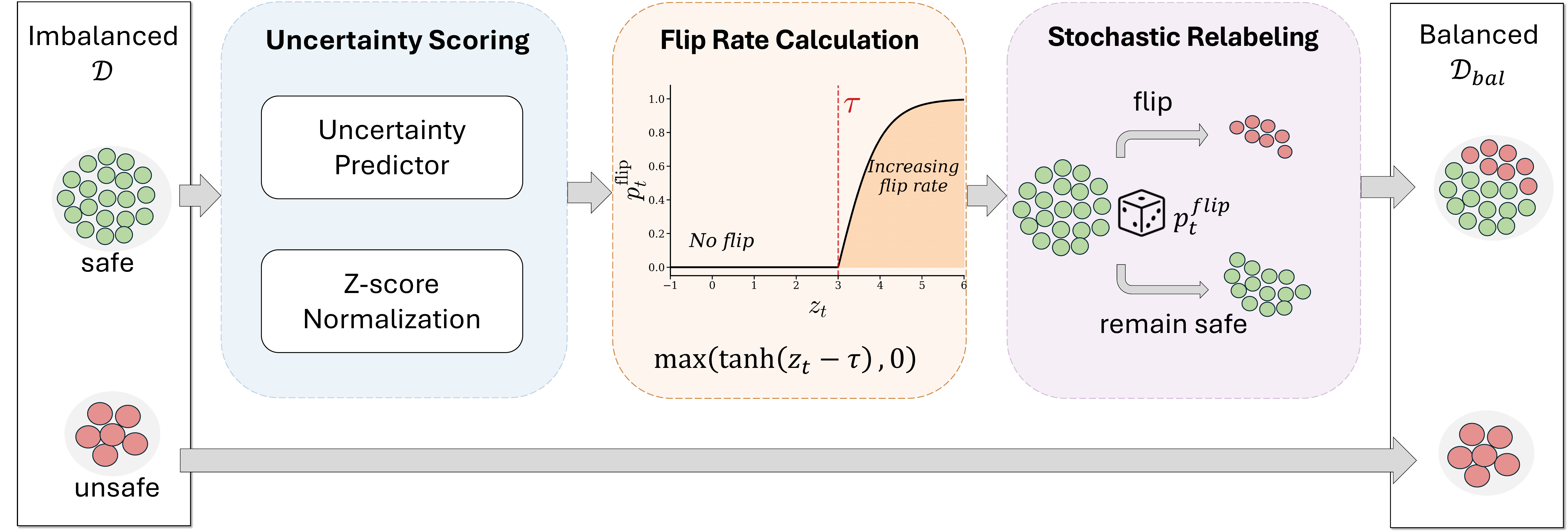}
    \caption{Overview of \ulnr (\textcircled{2}).}
    \label{fig:ulnr}
\end{figure}

Adapted from the original LNR~\cite{lnr}, \ulnr rebalances class labels in the training set using uncertainty information. Whereas LNR uses a classifier's predicted probabilities to guide relabeling, \ulnr uses predicted behavioral uncertainty. Specifically, it uses the output $\hat{u}_t$ from the uncertainty predictor to identify safe-labeled windows that exhibit unusually high behavioral uncertainty and to stochastically relabel them as unsafe. \ulnr operates in three steps, namely \textit{Uncertainty Scoring}, \textit{Flip-rate Calculation}, and \textit{Stochastic Relabeling}.

\noindent\emph{\textbf{Uncertainty Scoring.}} To characterize how unusual a sample is relative to typical safe samples in terms of uncertainty, we compute uncertainty scores for all safe windows and transform them into z-scores, which express each score as a signed distance from the mean in units of standard deviation. The z-score is calculated as $z_t = \frac{\hat{u}_t - \mu_{\mathcal{S}}}{\sigma_{\mathcal{S}} + \epsilon}$, where $\mu_{\mathcal{S}}$ and $\sigma_{\mathcal{S}}$ denote the mean and standard deviation of uncertainty scores among the set of safe training samples $\mathcal{S}$, and $\epsilon$ is a small constant for numerical stability.

A higher $z_t$ identifies a window with greater predicted uncertainty relative to the safe training group. We use this relative score to prioritize windows for relabeling.

\noindent\emph{\textbf{Flip-rate Calculation.}}
The z-scores are transformed into flip probabilities using a
shifted hyperbolic tangent as in Equation ~\ref{eq:flip_prob},
\begin{equation}
\label{eq:flip_prob}
    p_t^{\text{flip}} =
    \begin{cases}
        \max\!\big(\tanh(z_t - \tau),\; 0\big) & \text{if } l_t^s = 0, \\
        0 & \text{if } l_t^s = 1,
    \end{cases}
\end{equation}
where $\tau$ is a threshold parameter that controls relabeling aggressiveness. The $\tanh$ function provides a smooth, bounded mapping from z-scores to probabilities, while the $\max(\cdot, 0)$ operator ensures that samples with $z_t \leq \tau$ receive zero flip probability. Only safe-labeled samples are eligible for relabeling, increasing the number of minority-class labels.

\noindent\emph{\textbf{Stochastic Relabeling.}}
Each safe-labeled sample is independently relabeled by drawing from a Bernoulli trial parameterized by its flip probability as depicted in Equation ~\ref{eq:flip},
\begin{equation}
\label{eq:flip}
    \tilde{l}^s_t =
    \begin{cases}
        1 & \text{if } l^s_t = 0 \;\text{ and }\; \xi < p_t^{\text{flip}}, \\
        l^s_t & \text{otherwise},
    \end{cases}
    \quad \xi \sim \text{Uniform}(0, 1)
\end{equation}
The resulting label $\tilde l_t^s$ replaces $l_t^s$ in the rebalanced training set $\mathcal{D}_{bal}$. Eligible windows are relabeled probabilistically, with higher standardized uncertainty giving a higher flip probability. The telemetry windows themselves are unchanged.

\subsection{Safety Predictor}
\label{subsec:safety-predictor}
\begin{figure}
    \centering
    \includegraphics[width=0.35\linewidth]{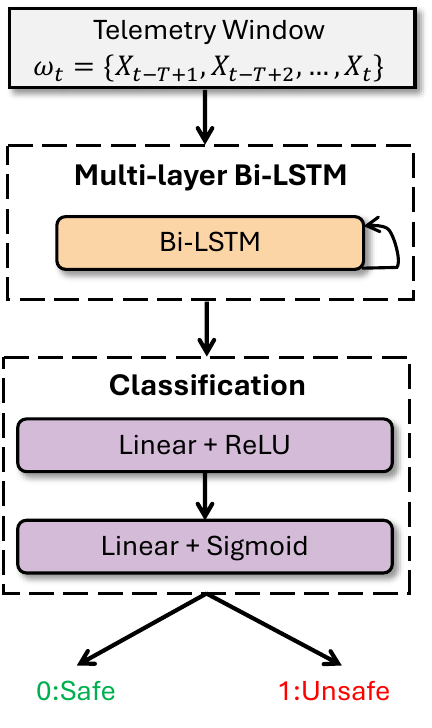}
    \caption{Overview of the Safety Predictor (\textcircled{3})}
    \label{fig:ms}
\end{figure}
The safety predictor learns from the raw telemetry windows and the labels produced by \ulnr. Each input is $\omega_t\in\mathbb{R}^{T\times C}$, with $T=25$ and $C=4$ on UAV, and $T=250$ and $C=9$ on ForEnt. Each channel is independently standardized. A Bi-LSTM encoder produces a window representation, and a classification head maps it to an estimated unsafe probability $\hat p_t^s\in[0,1]$. A decision threshold converts this probability into a binary safety prediction.

\rev{The relabeling procedure only modifies the training labels; it is architecture-agnostic and can be paired with any safety predictor. Predicting safety requires modeling how telemetry evolves over the input window, so we use a Bi-LSTM to capture temporal dependencies by processing the sequence forward and backward.}

\noindent\textbf{\textit{Bi-LSTM encoder}}. The encoder processes $\omega_t$ in both temporal directions~\cite{bilstm} and produces a window representation $h_t$.

\noindent\textbf{\textit{Classification}}. The classification head applies two linear layers separated by ReLU, followed by a sigmoid activation:
\begin{equation}
~\label{eq:safe_classification}
    \hat{p}_t^s = \sigma\!\big(W_2 \,  \text{ReLU}(W_1 \, h_t + b_1)  + b_2\big)
\end{equation}
The ReLU provides a nonlinear transformation of $h_t$, and the sigmoid $\sigma$ maps the final scalar to the unsafe-probability output.

\section{Experiment Design}
In Section ~\ref{subsec:rq}, we introduce three research questions (RQs), followed by dataset details in Section ~\ref{subsec:dataset}. We then present the baselines, evaluation metrics, and implementation details in Sections ~\ref{subsec:baselines}, ~\ref{subsec:metrics}, and ~\ref{subsec:implementation}, respectively.

\subsection{Research Questions}
\label{subsec:rq}

\begin{leftbar}
\noindent\textbf{RQ0 (Uncertainty and Safety)}: How is behavioral uncertainty related to safety, and how can it be used to improve safety prediction?
\end{leftbar}
\rev{We address RQ0 through two subquestions.}

\noindent\textit{RQ0.1 (Correlation): How is behavioral uncertainty correlated with safety?}
\rev{We investigate this relationship on both datasets by calculating the correlations of safety labels with uncertainty labels and with predicted uncertainty scores, using the original imbalanced data.}

\noindent\textit{RQ0.2 (Uncertainty Integration): How can behavioral uncertainty best be used to improve safety prediction?}
\rev{We compare three ways to use the same predicted uncertainty scores: adding the score to the safety predictor input (Early Fusion), adding it to the latent representation before classification (Late Fusion), and using it to relabel selected safe training windows as unsafe (\ulnr).}
\begin{leftbar}
\noindent\textbf{RQ1 (Effectiveness and Efficiency)}: How effective and efficient is \method in safety prediction compared with alternative predictors?
\end{leftbar}
This RQ evaluates \method's predictive performance and computational efficiency for safety monitoring. To this end, we conducted comprehensive experiments comparing \method against 14 baselines spanning classical machine learning, deep learning, and ensemble approaches. On both datasets, we assess effectiveness using precision, recall, F1, AUROC, and AUPRC, and efficiency using training time, model size, and inference latency. The plain Bi-LSTM uses the same safety-prediction architecture as \method, providing a control for the effect of relabeling.

\begin{leftbar}
\noindent\textbf{RQ2 (Ablation)}: How much do the key components contribute to \method's effectiveness, including the uncertainty predictor and the \ulnr?
\end{leftbar}
\rev{We address RQ2 through two subquestions.}

\noindent\textit{RQ2.1 (Uncertainty Predictor): How does the uncertainty predictor's design affect safety prediction?}
\rev{We conduct two types of ablations to examine the contributions of distributional statistics and the GatedMLP architecture to uncertainty-guided relabeling. One uses individual statistics directly as uncertainty scores, while the other replaces GatedMLP with alternative predictor architectures. Both keep \ulnr and the downstream Bi-LSTM unchanged.}

\noindent\textit{Direct statistical scores.} \rev{To examine whether extracting distributional statistics alone explains the improvement, we use individual standard deviations directly as uncertainty scores. For each dataset, we select the two standard deviations with the highest absolute Spearman correlations with the uncertainty labels, using only the windows used to train the uncertainty predictor. These are the heading and $y$-coordinate standard deviations on UAV, and the accelerometer channel 0 and gyroscope channel 1 standard deviations on ForEnt. We call them heading variability, $y$-position variability, acceleration variability, and angular-rate variability, respectively. Each score replaces the GatedMLP output, while \ulnr and the downstream safety predictor remain unchanged.}

\noindent\textit{Alternative predictor architectures.} This ablation evaluates the choice of GatedMLP for learning uncertainty scores. We replace it with a plain MLP trained on the same distributional statistics, and with Bi-RNN, Bi-GRU, and Bi-LSTM predictors trained on raw telemetry windows. The comparison with the plain MLP tests the gated architecture with the input features fixed. The sequential alternatives evaluate different architectures and input representations.

\noindent\textit{RQ2.2 (Uncertainty-Guided Relabeling): How effective is \ulnr for rebalancing the training data?}
We vary $\tau\in\{0.5,1.0,1.5,2.0,2.5,3.0,3.5,4.0\}$ to assess sensitivity to the amount of relabeling. We also compare \ulnr with 14 alternative rebalancing methods. Section~\ref{subsec:baselines} introduces these methods, and Section~\ref{subsec:implementation} gives their sampling configurations.

\subsection{Datasets}
\label{subsec:dataset}
\rev{We require datasets that meet three stringent criteria: (1) the dataset must support safety prediction and provide safety labels or annotations from which these labels can be derived, (2) the dataset must record the operation of a CPS, and (3) the dataset must include telemetry that reflects inherent uncertainty in system behavior. Based on these criteria, we could identify the UAV dataset of Khatiri et al.~\cite{superialist} and the ForEnt quadruped robot dataset~\cite{forent}.}

\noindent\textbf{UAV Dataset.}
The UAV benchmark of Khatiri et al.~\cite{superialist} provides collision-related safety labels and expert behavioral-uncertainty annotations for obstacle-avoidance flights. \rev{A window is labeled unsafe if the UAV comes within 1.5~m of any obstacle at any time step in that window.} It comprises 1,498 flights totaling approximately 53 hours and 39 minutes.

We generated the flights using Surrealist~\cite{surrealist}, an automated, simulation-based test case generation tool for UAVs. Surrealist generates challenging environments by iteratively introducing static obstacles into PX4 autopilot missions, adjusting their size, position, and orientation to create scenarios in which the PX4-Avoidance collision-prevention system struggles to identify a safe path. The tool also enables multiple simulations per test case to account for non-deterministic behavior. The resulting dataset covers a range of flight conditions, including safe, unsafe, certain, and uncertain behaviors, as illustrated in Figure \ref{fig:flight-types}.

We segment each UAV flight into fixed-length windows of 25 time points. \rev{We split UAV windows chronologically into training, validation, and test sets at an 8:1:1 ratio. We use the earliest windows for training, the following windows for validation, and the latest windows for testing, following common practice in time series evaluation~\cite{bergmeir2012crossvalidation,xu2021digital,xu2024pretrain}. The safe-to-unsafe ratios are approximately $76{:}1$, $8{:}1$, and $13{:}1$, respectively. All methods use the same partitions. Relabeling and rebalancing are applied only to training data.} The dataset contains 77,254 windows, with 61,803, 7,725, and 7,726 windows in the training, validation, and test splits, respectively.

\rev{\noindent\textbf{ForEnt.}
We use ForEnt as our second CPS case study~\cite{forent}. It records a teleoperated Unitree Go2 quadruped traversing forest terrain. We define unsafe operation using annotated entrapment intervals, during which the robot is stuck and unable to continue despite a movement command. A window is labeled unsafe if it overlaps an annotated entrapment interval.}

\rev{The dataset includes nine telemetry channels: three-axis acceleration, three-axis angular velocity, and roll, pitch, and yaw. Each 5-s window contains 250 time steps sampled at 50\,Hz, giving a $250\times9$ input.}

\rev{ForEnt also uses an 8:1:1 train/validation/test split. The training set contains 1,606 windows (1,224 safe and 382 unsafe), the validation set 163 (77 safe and 86 unsafe), and the test set 164 (142 safe and 22 unsafe). The unsafe proportions are 23.8\%, 52.8\%, and 13.4\%, with safe-to-unsafe ratios of $3.2{:}1$, $0.9{:}1$, and $6.5{:}1$, respectively. All methods use the same partitions.}

\rev{ForEnt does not provide behavioral uncertainty labels. Horizontal acceleration reflects changes in the robot's motion. We summarize the intensity of these changes within each window $\omega_t$ using mean horizontal-acceleration energy:
\[
e_t=\frac{1}{T}\sum_{k=1}^{T}\bigl(a_{x,t,k}^{2}+a_{y,t,k}^{2}\bigr),
\]
where $a_{x,t,k}$ and $a_{y,t,k}$ are the horizontal acceleration measurements and $T=250$.}

\rev{We use unusually high energy as a heuristic indicator of unstable motion and thus behavioral uncertainty. We treat scores above the 90th percentile among safe training windows as outliers and label the corresponding windows uncertain ($l_t^u=1$). All other windows are labeled certain ($l_t^u=0$).}

\rev{\noindent\textbf{Use of Data.} We partition each dataset once into training, validation, and test sets in temporal order. Both learning stages use this same partition. We train the uncertainty predictor only on training windows with binary uncertainty labels $l_t^u$. We then fix the uncertainty predictor and apply it to all windows in the same training split. \ulnr uses the predicted scores to update selected safety labels in that split. The safety predictor is trained from scratch on these same training windows with the updated safety labels. Both predictors use the same validation split for model selection, with uncertainty targets for the uncertainty predictor and original safety labels for the safety predictor. We evaluate the final safety predictor on the test split using its original safety labels. Neither validation nor test data are relabeled or rebalanced.}

\subsection{Baselines}
\label{subsec:baselines}
We use two sets of baselines. RQ1 compares the complete \method pipeline with 14 alternative safety predictors. RQ2.2 keeps the downstream Bi-LSTM fixed and compares the balancing methods used to prepare its training data.

\noindent\textbf{Classical ML.}
We include \textit{Logistic Regression (LR)}~\cite{cox1958regression}, \textit{Support Vector Machine (SVM)}~\cite{steinwart2008support}, \textit{Decision Tree (DT)}~\cite{quinlan1986induction}, \textit{K-Nearest Neighbors (KNN)}~\cite{cover1967nearest}, and \textit{Multi-Layer Perceptron (MLP)}~\cite{rumelhart1986learning} as standard classification baselines.

\noindent\textbf{Deep Learning.} The sequence-model baselines are CNN~\cite{lecun2002gradient}, Bi-LSTM~\cite{huang2015bidirectional}, Bi-GRU~\cite{cho2014learning}, and Transformer (TF)~\cite{vaswani2017attention}. We also evaluate TFT~\cite{lim2021temporal}, which combines recurrent processing with attention and has been used for safety-metric forecasting~\cite{sharifi2025system}, and a TimeMoE-based model using a mixture-of-experts architecture~\cite{timemoe}. Superialist~\cite{superialist} supplies a baseline developed for the UAV benchmark. Because its original evaluation is at the flight level, we adapt it to the window-level task used here.

\noindent\textbf{Ensemble Methods.}
We include \textit{Random Forest (RF)}~\cite{breiman2001random} and \textit{Gradient Boosting (GB)}~\cite{friedman2001greedy} as representative ensemble baselines. \textit{RF } builds an ensemble of DTs, while \textit{GB} constructs the ensemble sequentially, with each base classifier correcting the residual error of its predecessor.

\noindent\textbf{Rebalancing Methods.} \rev{For the \ulnr ablation, we compare 14 alternatives covering oversampling, undersampling, and changes to the training objective. Rebalancing is applied only to training data. SMOTE, ADASYN, Borderline-SMOTE, RUS, OSS, and CC treat each complete telemetry window as one sample. The resulting inputs are reshaped to the original window dimensions before fitting the safety predictor.}

\rev{SMOTE~\cite{chawla2002smote} creates synthetic unsafe windows by interpolating between an unsafe window and one of its nearest unsafe neighbors. ADASYN~\cite{he2008adasyn} allocates more synthetic windows to unsafe examples whose neighborhoods contain a larger proportion of safe examples. Borderline-SMOTE~\cite{han2005borderline} restricts SMOTE generation to unsafe examples near the class boundary, identified by the safe and unsafe examples in their neighborhoods. T-SMOTE~\cite{zhao2022t} uses temporal information to generate unsafe examples near the class boundary, synthesizes further examples from them, and selects from both sets through weighted sampling. RSG~\cite{wang2021rsg} generates features for unsafe examples by transferring variations learned within the safe class while filtering out information specific to that class.}

\rev{ReMix~\cite{chou2020remix} separates the input and label mixing weights so that mixed safe--unsafe examples can receive greater label weight for the unsafe class. SelMix~\cite{ramasubramanian2024selective} uses validation feedback to select which class pairs to mix in feature space during fine-tuning toward the target performance metric.}

\rev{RUS randomly discards safe windows until the prescribed ratio of safe to unsafe samples is reached. OSS~\cite{batista2000applying} reduces redundant safe windows through condensed nearest-neighbor selection, then removes the safe member of each mutually nearest safe--unsafe pair (a Tomek link). CC clusters safe windows using k-means and replaces each cluster with its centroid~\cite{macqueen1967multivariate}.}

\rev{Class Weighting (CW) weights the training loss by the inverse effective number of samples in each class~\cite{cui2019classbalanced}. LDAM-DRW~\cite{cao2019learning} gives the unsafe class a larger required classification margin and postpones increased weighting of that class until later training. Gaussian Clouded Logit adjustment (GCL)~\cite{li2022long} applies Gaussian perturbations to class logits with larger amplitudes for less frequent classes to counter the safe class's dominance. MiSLAS~\cite{zhong2021improving} learns representations with mixup and then trains the classifier using stronger label smoothing for the frequent safe class than for the unsafe class.}

\subsection{Evaluation Metrics and Statistical Testing}
\label{subsec:metrics}
\noindent\textit{Correlation.} We measure the correlations of safety labels with uncertainty labels and with predicted uncertainty scores on the original imbalanced datasets. We use the $\phi$ coefficient to measure the association between uncertainty and safety labels because both are binary. For continuous predicted uncertainty scores and binary safety labels, we use point-biserial correlation~\cite{crocker1986introduction,pb1,pb2}. The coefficients range from $-1$ to $1$, with values closer to zero indicating a weaker association.

\noindent\textit{Effectiveness.} We evaluate \method and all baselines using precision, recall, F1, AUROC, and AUPRC~\cite{hossin2015review}, as these are standard metrics for classification tasks, particularly under class imbalance. Precision, recall, and F1 are computed from the numbers of true positives (TP), true negatives (TN), false positives (FP), and false negatives (FN). In our context, TP, TN, FP, and FN denote unsafe windows correctly predicted as unsafe, safe windows correctly predicted as safe, safe windows incorrectly predicted as unsafe, and unsafe windows incorrectly predicted as safe, respectively. 

\rev{To compute AUROC, we vary the threshold on predicted unsafe scores and plot the true positive rate $\mathrm{TP}/(\mathrm{TP}+\mathrm{FN})$ against the false positive rate $\mathrm{FP}/(\mathrm{FP}+\mathrm{TN})$. AUROC is the area under this ROC curve. For AUPRC, we compute precision and recall at these thresholds and use average precision, $\mathrm{AUPRC}=\sum_n(R_n-R_{n-1})P_n$. Here, $P_n$ and $R_n$ are precision and recall as the threshold is lowered, with $R_0=0$. Each precision value is weighted by the corresponding increase in recall.}

\noindent\textit{Efficiency.} We report training time in seconds, inference latency in milliseconds per window, and model size. For neural sequence models, size is the parameter count. For classical models, it is the stored model size in KB. These size measures are reported separately by unit and should not be compared directly.

\noindent\textit{Statistical Analysis.} \rev{We repeat each experiment 30 times using different random seeds and report the mean and standard deviation of the results. We use paired Wilcoxon tests to assess whether \method achieves higher F1, AUROC, and AUPRC than each baseline. In the tables, \textsuperscript{*} denotes $p<0.05$. We report the Vargha--Delaney effect size $\hat{A}_{12}$~\cite{st}, the probability that a reference score exceeds a comparator score, counting ties as half. Following standard guidelines, we classify its magnitude using $A^\star=\max(\hat{A}_{12},1-\hat{A}_{12})$: (N) negligible for $A^\star<0.56$, (S) small for $0.56\leq A^\star<0.64$, (M) medium for $0.64\leq A^\star<0.71$, and (L) large for $A^\star\geq0.71$. Effect letters describe magnitude independently of significance and direction.}

\subsection{Implementation Details}
\label{subsec:implementation}

\rev{For each dataset, we select model hyperparameters using held-out validation data. Each prediction model receives the same budget of candidate configurations. Table~\ref{tab:dataset_hyperparameters} reports selected parameter values that differ from the implementation defaults. Other parameters retain those defaults. Our repository provides the candidate configurations, default values, and complete selected configurations.}

\rev{In Table~\ref{tab:dataset_hyperparameters}, \texttt{dim\_p} denotes GatedMLP's projection dimension $p$, and \texttt{dim\_hidden} denotes its intermediate hidden dimension $e$ (Section~\ref{subsec:uncertainty-predictor}). For the other models, \texttt{dim\_hidden} specifies the dimensions of their hidden representations. \texttt{dim\_model} specifies the Transformer representation dimension, and \texttt{dim\_channels} lists the number of output channels in the CNN layers.}

\begin{table*}[t]
\centering
\color{black}
\captionsetup{textfont={color=black},labelfont={bf,color=black}}
\caption{Tuned model hyperparameters. The \method row lists uncertainty-predictor settings. Bi-LSTM settings apply to both \method and the plain Bi-LSTM. Only selected values that differ from implementation defaults are listed. Default values and full configurations are provided in our repository.}
\label{tab:dataset_hyperparameters}
\begingroup
\footnotesize
\setlength{\tabcolsep}{4pt}
\renewcommand{\arraystretch}{1.15}
\begin{tabular}{@{}p{0.14\textwidth}p{0.40\textwidth}p{0.40\textwidth}@{}}
\toprule
\textbf{Method} & \textbf{UAV} & \textbf{ForEnt} \\
\midrule
\method & \texttt{dim\_p=64}; \texttt{dim\_hidden=128} & \texttt{dim\_p=64}; \texttt{dim\_hidden=128} \\
Bi-LSTM & \texttt{dim\_hidden=64}; \texttt{num\_layers=3} & \texttt{dim\_hidden=128}; \texttt{num\_layers=1} \\
\addlinespace[2pt]
CNN & \texttt{dim\_channels=[32,64,64]};\newline \texttt{kernel\_sizes=[5,5,3]} & \texttt{dim\_channels=[32,64,64]};\newline \texttt{kernel\_size=5} \\
\addlinespace[2pt]
Bi-GRU & \texttt{dim\_hidden=64}; \texttt{num\_layers=3} & \texttt{dim\_hidden=64}; \texttt{num\_layers=3} \\
\addlinespace[2pt]
Transformer & \texttt{dim\_model=64}; \texttt{num\_layers=2};\newline \texttt{num\_heads=4} & \texttt{dim\_model=64}; \texttt{num\_layers=2};\newline \texttt{num\_heads=4} \\
\addlinespace[2pt]
TFT & \texttt{dim\_hidden=64}; \texttt{num\_heads=4} & \texttt{dim\_hidden=64}; \texttt{num\_heads=4} \\
\addlinespace[2pt]
TimeMoE & \texttt{dim\_hidden=64}; \texttt{num\_experts=4};\newline \texttt{num\_expert\_layers=3} & \texttt{dim\_hidden=64}; \texttt{num\_experts=4};\newline \texttt{num\_expert\_layers=3} \\
\addlinespace[2pt]
Superialist & \texttt{threshold=1.5} & \texttt{threshold=1.5} \\
Logistic reg. & \texttt{C=10}; \texttt{solver=liblinear} & \texttt{C=10}; \texttt{solver=liblinear} \\
\addlinespace[2pt]
Decision tree & \texttt{max\_depth=15}; \texttt{min\_samples\_leaf=2} & \texttt{max\_depth=15}; \texttt{min\_samples\_leaf=2} \\
\addlinespace[2pt]
Random forest & \texttt{n\_estimators=50} & \texttt{n\_estimators=200}; \texttt{max\_depth=3};\newline \texttt{min\_samples\_split=5} \\
\addlinespace[2pt]
SVM & \texttt{C=10} & \texttt{C=10} \\
\addlinespace[2pt]
MLP & \texttt{dim\_hidden=(128,64)}; \texttt{alpha=0.01} & \texttt{dim\_hidden=(64,32)}; \texttt{alpha=1.0} \\
\addlinespace[2pt]
KNN & \texttt{n\_neighbors=7}; \texttt{weights=distance};\newline \texttt{metric=manhattan} & \texttt{n\_neighbors=7}; \texttt{weights=distance};\newline \texttt{metric=manhattan} \\
\addlinespace[2pt]
Grad.\ boosting & \texttt{n\_estimators=200}; \texttt{max\_depth=7};\newline \texttt{min\_samples\_split=5} & \texttt{n\_estimators=100}; \texttt{max\_depth=5};\newline \texttt{min\_samples\_split=5} \\
\bottomrule
\end{tabular}
\endgroup
\end{table*}

\rev{We select the flip threshold $\tau$ on the held-out validation dataset by maximizing the safety predictor's AUROC. Ties are broken first by AUPRC and then by F1. The selected values are $\tau=3.0$ for UAV and $\tau=0.5$ for ForEnt.}

\rev{For SMOTE, Borderline-SMOTE, RSG, T-SMOTE, RUS, and CC, we use the recommended $1{:}1$ ratio of safe to unsafe training samples on both datasets~\cite{chawla2002smote,han2005borderline,wang2021rsg,zhao2022t}. The final ratios of ADASYN and OSS are determined by their sampling procedures~\cite{he2008adasyn,batista2000applying}. Specifically, ADASYN targets class balance and calculates how many synthetic unsafe examples to generate from the local class distribution, while OSS removes redundant safe examples and safe members of mutually nearest safe--unsafe pairs. ReMix mixes inputs with labels weighted toward the unsafe class, while SelMix uses validation feedback to select class pairs for feature mixing. LDAM-DRW adjusts classification margins and class weights, GCL applies class-dependent perturbations to logits, and MiSLAS combines mixup with class-dependent label smoothing. CW assigns separate loss weights to safe and unsafe windows. These six methods do not use a target sampling ratio.}

\rev{We use recommended parameter values in their original paper and select parameters requiring tuning on the held-out validation dataset. On both datasets, we use \texttt{n\_neighbors=1} for OSS~\cite{batista2000applying}, $\alpha=0.4$ for ReMix~\cite{chou2020remix}, and $\alpha=0.3$ for SelMix~\cite{ramasubramanian2024selective}. For LDAM-DRW~\cite{cao2019learning}, we use \texttt{max\_m=0.5} and $\beta=0.9999$, with class reweighting starting at 80\% of training epochs. For GCL, we use $\beta=0.9999$ and \texttt{noise\_mul=0} on UAV, and $\beta=0.99$ and \texttt{noise\_mul=1.5} on ForEnt. For MiSLAS, we use label smoothing $\epsilon=0.1$.}

\rev{For CW, we use the class-balanced weighting of Cui et al.~\cite{cui2019classbalanced}, with $\beta=0.99$ on both datasets. The unsafe/safe weights are approximately $1/1$ on UAV and $1.011/0.989$ on ForEnt. We provide the full search grids in our repository.}

We conducted all experiments on a workstation equipped with an Intel Xeon w9-3495X processor and 128 GB of RAM. The implementation uses Python 3.10 and PyTorch 2.9~\cite{pytorch}. We will release our code and data publicly for replication upon acceptance.

\section{Results and Analysis}

This section presents results for all RQs on the UAV and ForEnt datasets. Results are reported as mean $\pm$ standard deviation unless otherwise indicated. Statistical tests and effect sizes follow Section~\ref{subsec:metrics}.

\subsection{RQ0: Uncertainty and Safety}

\subsubsection{RQ0.1: Correlation between Uncertainty and Safety}
\label{subsubsec:rq0-correlation}

\rev{We first compare the correlations of uncertainty labels and predicted uncertainty scores with safety on the original imbalanced datasets. We use the $\phi$ coefficient between uncertainty and safety labels because both are binary, and point-biserial correlation between predicted uncertainty scores and safety labels because the scores are continuous and the labels are binary. On UAV, uncertainty and safety labels are strongly correlated ($0.885$), but the predicted scores have a weaker correlation with safety ($0.449\pm0.023$). On ForEnt, the correlation between uncertainty and safety labels is moderate ($0.510$), and, unlike UAV, the predicted scores show a similar correlation with safety ($0.517\pm0.003$).} In summary, on both datasets, predicted uncertainty scores are moderately correlated with safety. This indicates that predicted uncertainty can help predict safety, but it needs to be integrated into safety prediction effectively because the correlation is only moderate.

\subsubsection{RQ0.2: Uncertainty Integration}
\label{subsubsec:rq0-integration}

\rev{We next evaluate how to use this uncertainty information for safety prediction. We compare three approaches using the same predicted scores. Early Fusion adds the score to the safety predictor's input, and Late Fusion concatenates it with the learned representation before classification~\cite{10.1145/1101149.1101236,gadzicki2020early}. \ulnr uses the score to relabel training windows.}

\rev{We also include Plain, which uses the same Bi-LSTM architecture, telemetry inputs, and training procedure as the safety predictor in \method. It is trained on the original safety labels without uncertainty or rebalancing.}

\begin{table*}[t]
\centering
\color{black}
\captionsetup{textfont={color=black},labelfont={bf,color=black}}
\caption{Comparison of uncertainty integration strategies on UAV and ForEnt (RQ0.2). Values are mean $\pm$ standard deviation. The reference method is \ulnr. Bold numbers indicate the highest mean per dataset, including ties.}
\label{tab:rq1}
\resizebox{0.79\textwidth}{!}{
\begin{tabular}{l l ccccc}
\toprule
\textbf{Dataset} & \textbf{Strategy} & \textbf{Precision} & \textbf{Recall} & \textbf{F1} & \textbf{AUROC} & \textbf{AUPRC} \\
\midrule
\multirow{4}{*}{\textbf{UAV}} & \ulnr (ours) & 0.792$\pm$0.064 & \textbf{0.814$\pm$0.051} & \textbf{0.801$\pm$0.039} & \textbf{0.972$\pm$0.005} & \textbf{0.776$\pm$0.033} \\
 & Plain & \textbf{0.873$\pm$0.052} & 0.498$\pm$0.025 & 0.633$\pm$0.022\textsuperscript{*}\,(L) & 0.939$\pm$0.012\textsuperscript{*}\,(L) & 0.670$\pm$0.030\textsuperscript{*}\,(L) \\
 & Early Fusion & 0.851$\pm$0.030 & 0.520$\pm$0.017 & 0.645$\pm$0.014\textsuperscript{*}\,(L) & 0.958$\pm$0.010\textsuperscript{*}\,(L) & 0.722$\pm$0.030\textsuperscript{*}\,(L) \\
 & Late Fusion & 0.848$\pm$0.036 & 0.526$\pm$0.016 & 0.648$\pm$0.009\textsuperscript{*}\,(L) & 0.952$\pm$0.013\textsuperscript{*}\,(L) & 0.725$\pm$0.021\textsuperscript{*}\,(L) \\
\midrule
\multirow{4}{*}{\textbf{ForEnt}} & \ulnr (ours) & \textbf{0.408$\pm$0.071} & \textbf{0.776$\pm$0.152} & \textbf{0.522$\pm$0.045} & \textbf{0.806$\pm$0.050} & \textbf{0.323$\pm$0.045} \\
 & Plain & 0.370$\pm$0.148 & 0.674$\pm$0.151 & 0.450$\pm$0.076\textsuperscript{*}\,(L) & 0.686$\pm$0.069\textsuperscript{*}\,(L) & 0.279$\pm$0.112\textsuperscript{*}\,(L) \\
 & Early Fusion & 0.336$\pm$0.073 & 0.674$\pm$0.111 & 0.442$\pm$0.066\textsuperscript{*}\,(L) & 0.684$\pm$0.070\textsuperscript{*}\,(L) & 0.254$\pm$0.077\textsuperscript{*}\,(L) \\
 & Late Fusion & 0.324$\pm$0.080 & 0.686$\pm$0.095 & 0.429$\pm$0.073\textsuperscript{*}\,(L) & 0.666$\pm$0.081\textsuperscript{*}\,(L) & 0.240$\pm$0.057\textsuperscript{*}\,(L) \\
\bottomrule
\end{tabular}
}
\par\smallskip
\begin{minipage}{0.79\textwidth}\footnotesize \textsuperscript{*}$p<0.05$ for the reference method scoring higher. (N)/(S)/(M)/(L): negligible/small/medium/large $\hat{A}_{12}$ magnitude.
\end{minipage}
\end{table*}

\rev{Table~\ref{tab:rq1} shows that \ulnr achieves the highest mean F1, AUROC, and AUPRC on both datasets. On UAV, its F1 is $0.801\pm0.039$, compared with $0.633\pm0.022$ for Plain, $0.645\pm0.014$ for Early Fusion, and $0.648\pm0.009$ for Late Fusion. On ForEnt, the corresponding scores are $0.522\pm0.045$, $0.450\pm0.076$, $0.442\pm0.066$, and $0.429\pm0.073$.}

\rev{The mean AUROC and AUPRC of \ulnr are $0.972\pm0.005$ and $0.776\pm0.033$ on UAV, and $0.806\pm0.050$ and $0.323\pm0.045$ on ForEnt. In both UAV and ForEnt, its improvements in F1, AUROC, and AUPRC over all three alternatives are statistically significant ($p<0.001$), with large effect sizes ($\hat A_{12}\geq0.892$ in UAV and $\hat A_{12}\geq0.749$ in ForEnt)}. 

\begin{tcolorbox}
Predicted uncertainty scores are moderately correlated with safety on both datasets. Among the evaluated strategies, \ulnr is the most effective way to use uncertainty for safety prediction, achieving higher F1, AUROC, and AUPRC than Early Fusion and Late Fusion on both datasets.
\end{tcolorbox}

\subsection{RQ1: Safety Prediction Effectiveness and Efficiency}

\begin{table*}[t]
\centering
\color{black}
\captionsetup{textfont={color=black},labelfont={bf,color=black}}
\caption{Safety-prediction performance and efficiency on UAV and ForEnt. Values are mean $\pm$ standard deviation where available. Single values are point estimates. Model size is reported as parameter count (K) for neural sequence models and stored size (KB) for classical models. Training time is in seconds. Inference latency is in milliseconds per window. For \method, model size covers the safety predictor. The statistical reference is \method. Bold numbers indicate the highest mean per dataset, including ties.}
\label{tab:rq2}
\setlength{\tabcolsep}{2pt}
\resizebox{0.94\textwidth}{!}{
\begin{tabular}{l l ccccc rrr}
\toprule
\textbf{Dataset} & \textbf{Method} & \textbf{Precision} & \textbf{Recall} & \textbf{F1} & \textbf{AUROC} & \textbf{AUPRC} & \textbf{Train (s)} & \textbf{Size} & \textbf{Infer. (ms)} \\
\midrule
\multirow{15}{*}{\textbf{UAV}} & \method (ours) & 0.792$\pm$0.064 & \textbf{0.814$\pm$0.051} & \textbf{0.801$\pm$0.039} & \textbf{0.972$\pm$0.005} & \textbf{0.776$\pm$0.033} & 91.0$\pm$1.5 & 238.7K & 0.093 \\
  & Bi-LSTM & 0.873$\pm$0.052 & 0.498$\pm$0.025 & 0.633$\pm$0.022\textsuperscript{*}\,(L) & 0.939$\pm$0.012\textsuperscript{*}\,(L) & 0.670$\pm$0.030\textsuperscript{*}\,(L) & 69.9$\pm$1.0 & 238.7K & 0.093 \\
  & CNN & 0.855$\pm$0.036 & 0.513$\pm$0.019 & 0.640$\pm$0.013\textsuperscript{*}\,(L) & 0.950$\pm$0.007\textsuperscript{*}\,(L) & 0.668$\pm$0.021\textsuperscript{*}\,(L) & 54.7$\pm$1.1 & 25.8K & 0.0047 \\
  & Bi-GRU & 0.813$\pm$0.068 & 0.473$\pm$0.030 & 0.596$\pm$0.032\textsuperscript{*}\,(L) & 0.917$\pm$0.024\textsuperscript{*}\,(L) & 0.566$\pm$0.067\textsuperscript{*}\,(L) & 65.9$\pm$1.7 & 180.0K & 0.052 \\
  & TF & 0.394$\pm$0.264 & 0.410$\pm$0.210 & 0.350$\pm$0.176\textsuperscript{*}\,(L) & 0.726$\pm$0.127\textsuperscript{*}\,(L) & 0.301$\pm$0.174\textsuperscript{*}\,(L) & 130.1$\pm$20.3 & 102.4K & 0.032 \\
  & TFT & 0.827$\pm$0.078 & 0.484$\pm$0.030 & 0.608$\pm$0.029\textsuperscript{*}\,(L) & 0.903$\pm$0.063\textsuperscript{*}\,(L) & 0.601$\pm$0.054\textsuperscript{*}\,(L) & 85.9$\pm$1.2 & 106.7K & 0.035 \\
  & TimeMoE & 0.674$\pm$0.397 & 0.439$\pm$0.235 & 0.499$\pm$0.282\textsuperscript{*}\,(L) & 0.759$\pm$0.170\textsuperscript{*}\,(L) & 0.493$\pm$0.262\textsuperscript{*}\,(L) & 218.7$\pm$31.5 & 371.7K & 0.119 \\
  & Superialist & 0.084$\pm$0.011 & 0.441$\pm$0.216 & 0.136$\pm$0.009\textsuperscript{*}\,(L) & 0.475$\pm$0.017\textsuperscript{*}\,(L) & 0.066$\pm$0.004\textsuperscript{*}\,(L) & 351.6$\pm$18.2 & 4.1K & 0.046 \\
  & LR & 0.508$\pm$0.000 & 0.219$\pm$0.000 & 0.306$\pm$0.000\textsuperscript{*}\,(L) & 0.618$\pm$0.000\textsuperscript{*}\,(L) & 0.274$\pm$0.000\textsuperscript{*}\,(L) & 10.1$\pm$1.3 & 1.4 KB & 0.0002 \\
  & DT & 0.892$\pm$0.014 & 0.483$\pm$0.009 & 0.626$\pm$0.009\textsuperscript{*}\,(L) & 0.740$\pm$0.004\textsuperscript{*}\,(L) & 0.481$\pm$0.015\textsuperscript{*}\,(L) & 8.4$\pm$0.2 & 33 KB & 0.0001 \\
  & RF & \textbf{0.919$\pm$0.024} & 0.510$\pm$0.008 & 0.656$\pm$0.004\textsuperscript{*}\,(L) & 0.796$\pm$0.005\textsuperscript{*}\,(L) & 0.594$\pm$0.005\textsuperscript{*}\,(L) & 1.5$\pm$0.1 & 1822 KB & 0.0078 \\
  & SVM & 0.813$\pm$0.000 & 0.498$\pm$0.000 & 0.618$\pm$0.000\textsuperscript{*}\,(L) & 0.947$\pm$0.000\textsuperscript{*}\,(L) & 0.690$\pm$0.000\textsuperscript{*}\,(L) & 40.7$\pm$2.2 & 1144 KB & 0.109 \\
  & MLP & 0.896$\pm$0.047 & 0.522$\pm$0.024 & 0.658$\pm$0.008\textsuperscript{*}\,(L) & 0.946$\pm$0.007\textsuperscript{*}\,(L) & 0.715$\pm$0.023\textsuperscript{*}\,(L) & 52.0$\pm$21.8 & 338 KB & 0.011 \\
  & KNN & 0.811$\pm$0.000 & 0.538$\pm$0.000 & 0.647$\pm$0.000\textsuperscript{*}\,(L) & 0.764$\pm$0.000\textsuperscript{*}\,(L) & 0.548$\pm$0.000\textsuperscript{*}\,(L) & 2.2$\pm$0.0 & 30292 KB & 1.706 \\
  & GB & 0.896$\pm$0.048 & 0.509$\pm$0.014 & 0.648$\pm$0.004\textsuperscript{*}\,(L) & 0.920$\pm$0.014\textsuperscript{*}\,(L) & 0.668$\pm$0.020\textsuperscript{*}\,(L) & 1417.4$\pm$10.1 & 699 KB & 0.0033 \\
\midrule
\multirow{15}{*}{\textbf{ForEnt}} & \method (ours) & 0.408$\pm$0.071 & 0.776$\pm$0.152 & \textbf{0.522$\pm$0.045} & \textbf{0.806$\pm$0.050} & \textbf{0.323$\pm$0.045} & 56.5$\pm$10.3 & 150.6K & 0.646 \\
  & Bi-LSTM & 0.370$\pm$0.148 & 0.674$\pm$0.151 & 0.450$\pm$0.076\textsuperscript{*}\,(L) & 0.686$\pm$0.069\textsuperscript{*}\,(L) & 0.279$\pm$0.112\textsuperscript{*}\,(L) & 101.2$\pm$11.0 & 150.6K & 0.692 \\
  & CNN & \textbf{0.419$\pm$0.160} & 0.655$\pm$0.175 & 0.478$\pm$0.097\textsuperscript{*}\,(L) & 0.710$\pm$0.083\textsuperscript{*}\,(L) & 0.294$\pm$0.119\textsuperscript{*}\,(L) & 7.6$\pm$1.1 & 26.6K & 0.055 \\
  & Bi-GRU & 0.416$\pm$0.196 & 0.529$\pm$0.151 & 0.434$\pm$0.106\textsuperscript{*}\,(L) & 0.642$\pm$0.095\textsuperscript{*}\,(L) & 0.309$\pm$0.138\textsuperscript{*}\,(L) & 7.2$\pm$0.2 & 182.0K & 1.450 \\
  & TF & 0.374$\pm$0.177 & 0.748$\pm$0.168 & 0.460$\pm$0.121\textsuperscript{*}\,(L) & 0.714$\pm$0.146\textsuperscript{*}\,(L) & 0.308$\pm$0.166\textsuperscript{*}\,(L) & 20.2$\pm$1.3 & 102.7K & 0.530 \\
  & TFT & 0.357$\pm$0.155 & 0.676$\pm$0.149 & 0.432$\pm$0.093\textsuperscript{*}\,(L) & 0.677$\pm$0.095\textsuperscript{*}\,(L) & 0.259$\pm$0.113\textsuperscript{*}\,(L) & 13.5$\pm$1.6 & 107.0K & 1.925 \\
  & TimeMoE & 0.269$\pm$0.124 & 0.712$\pm$0.212 & 0.360$\pm$0.108\textsuperscript{*}\,(L) & 0.566$\pm$0.168\textsuperscript{*}\,(L) & 0.224$\pm$0.118\textsuperscript{*}\,(L) & 25.3$\pm$2.5 & 378.1K & 2.158 \\
  & Superialist & 0.366$\pm$0.229 & 0.718$\pm$0.343 & 0.347$\pm$0.035\textsuperscript{*}\,(L) & 0.656$\pm$0.056\textsuperscript{*}\,(L) & 0.295$\pm$0.108\textsuperscript{*}\,(L) & 33.1$\pm$2.9 & 4.1K & 0.477 \\
  & LR & 0.300$\pm$0.000 & 0.273$\pm$0.000 & 0.286$\pm$0.000\textsuperscript{*}\,(L) & 0.442$\pm$0.000\textsuperscript{*}\,(L) & 0.242$\pm$0.000\textsuperscript{*}\,(L) & 2.7$\pm$0.1 & 18 KB & 0.0038 \\
  & DT & 0.215$\pm$0.038 & 0.359$\pm$0.079 & 0.269$\pm$0.051\textsuperscript{*}\,(L) & 0.484$\pm$0.051\textsuperscript{*}\,(L) & 0.154$\pm$0.024\textsuperscript{*}\,(L) & 4.6$\pm$0.1 & 12 KB & 0.0016 \\
  & RF & 0.226$\pm$0.045 & 0.741$\pm$0.152 & 0.338$\pm$0.047\textsuperscript{*}\,(L) & 0.586$\pm$0.055\textsuperscript{*}\,(L) & 0.161$\pm$0.021\textsuperscript{*}\,(L) & 3.0$\pm$0.1 & 262 KB & 0.077 \\
  & SVM & 0.266$\pm$0.004 & 0.636$\pm$0.000 & 0.375$\pm$0.004\textsuperscript{*}\,(L) & 0.595$\pm$0.000\textsuperscript{*}\,(L) & 0.173$\pm$0.000\textsuperscript{*}\,(L) & 10.5$\pm$0.6 & 21365 KB & 1.452 \\
  & MLP & 0.137$\pm$0.010 & \textbf{0.979$\pm$0.066} & 0.239$\pm$0.010\textsuperscript{*}\,(L) & 0.317$\pm$0.085\textsuperscript{*}\,(L) & 0.103$\pm$0.019\textsuperscript{*}\,(L) & 1.9$\pm$0.1 & 4638 KB & 0.0063 \\
  & KNN & 0.151$\pm$0.000 & 0.364$\pm$0.000 & 0.213$\pm$0.000\textsuperscript{*}\,(L) & 0.460$\pm$0.000\textsuperscript{*}\,(L) & 0.119$\pm$0.000\textsuperscript{*}\,(L) & 1.8$\pm$0.0 & 14128 KB & 0.842 \\
  & GB & 0.282$\pm$0.015 & 0.724$\pm$0.073 & 0.405$\pm$0.015\textsuperscript{*}\,(L) & 0.696$\pm$0.008\textsuperscript{*}\,(L) & 0.213$\pm$0.004\textsuperscript{*}\,(L) & 170.2$\pm$6.0 & 388 KB & 0.0059 \\
\bottomrule
\end{tabular}
}
\par\smallskip
\begin{minipage}{0.94\textwidth}\footnotesize \textsuperscript{*}$p<0.05$ for the reference method scoring higher. (N)/(S)/(M)/(L): negligible/small/medium/large $\hat{A}_{12}$ magnitude.
\end{minipage}
\end{table*}

We evaluate whether uncertainty-guided relabeling improves safety prediction across model families and what computational overhead it introduces. Table~\ref{tab:rq2} compares \method against 14 baselines on UAV and ForEnt in terms of predictive performance, training time, model size, and inference latency. The Bi-LSTM baseline corresponds to the Plain predictor from RQ0.2.

\noindent\textbf{Effectiveness.} \rev{\method achieves the highest mean F1, AUROC, and AUPRC among the 14 baselines on both datasets. On UAV, its F1 improvements over all baselines are significant ($p<0.001$), with $\hat A_{12}$ ranging from $0.968$ to $1.000$. Its UAV AUROC improvements are also significant ($p<0.001$).}

\rev{On UAV, \method achieves F1 of $0.801\pm0.039$, compared with $0.658\pm0.008$ for MLP, the strongest baseline, a gain of 14.3 percentage points. Its recall is higher than MLP's ($0.814$ versus $0.522$), while precision decreases from $0.896$ to $0.792$. The recall improvement yields a higher F1 despite this precision drop. \method also achieves the highest AUROC and AUPRC ($0.972\pm0.005$ and $0.776\pm0.033$, respectively). All AUPRC gains are significant with large effects ($p<0.001$, $\hat A_{12}\geq0.934$).}

\rev{On ForEnt, \method attains an F1 of $0.522\pm0.045$, outperforming CNN---the strongest baseline---at $0.478\pm0.097$ (+4.4 percentage points). The gain is driven by higher recall ($0.776$ vs.~$0.655$), while precision is comparable ($0.408$ vs.~$0.419$). \method also achieves the best AUROC and AUPRC ($0.806\pm0.050$ and $0.323\pm0.045$). Among baselines, Transformer yields the highest AUROC ($0.714\pm0.146$), and Bi-GRU yields the highest AUPRC ($0.309\pm0.138$). Compared with the strongest baseline for each metric, \method improves F1 by 4.4 percentage points over CNN, AUROC by 9.2 percentage points over Transformer, and AUPRC by 1.4 percentage points over Bi-GRU. The F1, AUROC, and AUPRC improvements over all baselines are significant, with large effects. The best baseline varies across metrics, so the reported gains are conservative: comparing with any single baseline would yield larger gains on some metrics. These results show that \method delivers the best overall safety-prediction performance, rather than performing well on only one metric.}

\rev{The F1 gain is substantially larger on UAV than on ForEnt (14.3 vs.~4.4 percentage points). A plausible explanation is the heavier class imbalance in UAV, where the safe-to-unsafe ratio is about $76{:}1$, compared with $3.2{:}1$ on ForEnt. Nevertheless, the ForEnt results indicate that \method remains effective under more moderate imbalance.}

\noindent\textbf{Efficiency.} \rev{Because the uncertainty predictor and relabeling are used only during training, \method deploys the same runtime architecture as the plain Bi-LSTM. On UAV, the safety predictor has 238.7K parameters and an inference latency of $0.093$\,ms per window; on ForEnt, it has 150.6K parameters and a latency of $0.646$\,ms. The plain Bi-LSTM has the same UAV latency ($0.093$\,ms)and slightly higher ForEnt latency ($0.692$\,ms).}

\rev{Training \method takes $91.0\pm1.5$\,s compared with $69.9\pm1.0$\,s for Plain on UAV, representing approximately 30\% additional training time. On ForEnt, \method takes $56.5\pm10.3$\,s versus $101.2\pm11.0$\,s for Plain, a reduction of approximately 44\%. Despite the additional uncertainty-estimation and relabeling stages, \method does not consistently require longer training time than Plain Bi-LSTM across the two datasets, and incurs a moderate overhead on UAV, while achieving a shorter measured training time on ForEnt.}

\rev{At inference time, \method retains the same architecture and params as Plain Bi-LSTM, with identical latency on UAV ($0.093$\,ms) and comparable latency for ForEnt ($0.646$\,ms versus $0.692$\,ms). Because uncertainty-guided relabeling occurs during training, \method's improved predictive performance adds no inference overhead.}

\begin{tcolorbox}
\method achieves the highest F1, AUROC, and AUPRC among the 14 baselines on both datasets, with practical training costs and low inference latency.
\end{tcolorbox}

\subsection{RQ2: Ablation Study}

\begin{table*}[t]
\centering
\color{black}
\captionsetup{textfont={color=black},labelfont={bf,color=black}}
\caption{Comparison of uncertainty scoring methods (RQ2.1). Downstream prediction metrics are mean $\pm$ standard deviation. Variability heuristics use within-window standard deviations. Their inputs and selection are described in Section~\ref{subsec:rq}. Selection uses only training data. The reference method is GatedMLP. Bold numbers indicate the highest mean per dataset, including ties.}
\label{tab:rq3_dt_ablation}
\label{tab:uncertainty_estimators}
\resizebox{0.93\textwidth}{!}{
\begin{tabular}{l l ccccc}
\toprule
\textbf{Dataset} & \textbf{Uncertainty scoring} & \textbf{Precision} & \textbf{Recall} & \textbf{F1} & \textbf{AUROC} & \textbf{AUPRC} \\
\midrule
\multirow{7}{*}{\textbf{UAV}} & GatedMLP (ours) & 0.792$\pm$0.064 & \textbf{0.814$\pm$0.051} & \textbf{0.801$\pm$0.039} & \textbf{0.972$\pm$0.005} & \textbf{0.776$\pm$0.033} \\
 & Plain MLP & 0.713$\pm$0.058 & 0.757$\pm$0.053 & 0.732$\pm$0.042\textsuperscript{*}\,(L) & 0.952$\pm$0.012\textsuperscript{*}\,(L) & 0.690$\pm$0.044\textsuperscript{*}\,(L) \\
 & Bi-LSTM & 0.753$\pm$0.090 & 0.690$\pm$0.126 & 0.707$\pm$0.066\textsuperscript{*}\,(L) & 0.966$\pm$0.007\textsuperscript{*}\,(L) & 0.736$\pm$0.070\textsuperscript{*}\,(M) \\
 & Bi-GRU & 0.727$\pm$0.099 & 0.688$\pm$0.087 & 0.697$\pm$0.051\textsuperscript{*}\,(L) & 0.955$\pm$0.010\textsuperscript{*}\,(L) & 0.741$\pm$0.042\textsuperscript{*}\,(L) \\
 & Bi-RNN & 0.769$\pm$0.080 & 0.652$\pm$0.116 & 0.695$\pm$0.068\textsuperscript{*}\,(L) & 0.964$\pm$0.012\textsuperscript{*}\,(M) & 0.724$\pm$0.044\textsuperscript{*}\,(L) \\
 & Heading variability (heuristic) & \textbf{0.891$\pm$0.116} & 0.417$\pm$0.046 & 0.562$\pm$0.041\textsuperscript{*}\,(L) & 0.909$\pm$0.012\textsuperscript{*}\,(L) & 0.592$\pm$0.030\textsuperscript{*}\,(L) \\
 & $y$-position variability (heuristic) & 0.848$\pm$0.064 & 0.508$\pm$0.046 & 0.631$\pm$0.021\textsuperscript{*}\,(L) & 0.936$\pm$0.017\textsuperscript{*}\,(L) & 0.676$\pm$0.041\textsuperscript{*}\,(L) \\
\midrule
\multirow{7}{*}{\textbf{ForEnt}} & GatedMLP (ours) & 0.408$\pm$0.071 & \textbf{0.776$\pm$0.152} & \textbf{0.522$\pm$0.045} & \textbf{0.806$\pm$0.050} & \textbf{0.323$\pm$0.045} \\
 & Plain MLP & 0.415$\pm$0.077 & 0.680$\pm$0.113 & 0.504$\pm$0.045\,(S) & 0.768$\pm$0.050\textsuperscript{*}\,(M) & 0.298$\pm$0.038\,(N) \\
 & Bi-LSTM & 0.370$\pm$0.082 & 0.670$\pm$0.104 & 0.467$\pm$0.070\textsuperscript{*}\,(L) & 0.711$\pm$0.067\textsuperscript{*}\,(L) & 0.275$\pm$0.076\textsuperscript{*}\,(M) \\
 & Bi-GRU & \textbf{0.416$\pm$0.120} & 0.683$\pm$0.091 & 0.503$\pm$0.068\textsuperscript{*}\,(L) & 0.726$\pm$0.049\textsuperscript{*}\,(L) & 0.307$\pm$0.098\textsuperscript{*}\,(L) \\
 & Bi-RNN & 0.371$\pm$0.109 & 0.680$\pm$0.113 & 0.465$\pm$0.062\textsuperscript{*}\,(L) & 0.709$\pm$0.066\textsuperscript{*}\,(L) & 0.286$\pm$0.078\textsuperscript{*}\,(L) \\
 & Acceleration variability (heuristic) & 0.391$\pm$0.060 & 0.712$\pm$0.087 & 0.498$\pm$0.048\textsuperscript{*}\,(L) & 0.741$\pm$0.039\textsuperscript{*}\,(L) & 0.309$\pm$0.065\textsuperscript{*}\,(L) \\
 & Angular-rate variability (heuristic) & 0.400$\pm$0.078 & 0.706$\pm$0.084 & 0.503$\pm$0.059\textsuperscript{*}\,(L) & 0.724$\pm$0.046\textsuperscript{*}\,(L) & 0.307$\pm$0.079\textsuperscript{*}\,(L) \\
\bottomrule
\end{tabular}
}
\par\smallskip
\begin{minipage}{0.93\textwidth}\footnotesize \textsuperscript{*}$p<0.05$ for the reference method scoring higher. (N)/(S)/(M)/(L): negligible/small/medium/large $\hat{A}_{12}$ magnitude.
\end{minipage}
\end{table*}

\subsubsection{RQ2.1: Ablation of Uncertainty Predictor}
\label{subsubsec:uncertainty-ablation}
\rev{The uncertainty predictor combines a statistical representation of each window with a learned mapping to uncertainty. We examine the contribution of each design choice through two ablations: using selected statistics directly as uncertainty scores and replacing GatedMLP with alternative predictor architectures. Table~\ref{tab:rq3_dt_ablation} reports the downstream safety-prediction results using the same \ulnr procedure and Bi-LSTM safety-predictor architecture across all comparisons.}

\noindent\textit{Direct statistical scores.} \rev{On UAV, GatedMLP achieves F1 of $0.801\pm0.039$, compared with $0.562\pm0.041$ for heading variability and $0.631\pm0.021$ for $y$-position variability. It also achieves higher AUROC and AUPRC than both heuristics. All three metrics differ significantly ($p<0.001$), with large effects ($\hat A_{12}\geq0.977$).}

\rev{On ForEnt, GatedMLP achieves higher recall than acceleration and angular-rate variability ($0.776$ versus $0.712$ and $0.706$), as well as higher AUROC ($0.806$ versus $0.741$ and $0.724$). Its F1 is $0.522$, compared with $0.498$ and $0.503$, and its AUPRC is $0.323$, compared with $0.309$ and $0.307$. The improvements in AUROC are statistically significant ($p<0.001$) with large effect sizes ($\hat A_{12}\geq0.843$). The F1 improvement is significant only against acceleration variability ($p=0.010$, $\hat{A}_{12}=0.612$), while neither AUPRC improvement is significant ($p\geq0.118$, $\hat{A}_{12}=0.611$--$0.620$).}

\rev{These results indicate that the improvement is not due to extracting distributional statistics alone:  GatedMLP also contributes to safety prediction.}

\noindent\textit{Alternative predictor architectures.} Replacing GatedMLP with a plain MLP trained on the same distributional features reduces UAV F1 from $0.801\pm0.039$ to $0.732\pm0.042$ ($p=1.4\times10^{-6}$, $\hat A_{12}=0.929$). \rev{On ForEnt, GatedMLP achieves F1 of $0.522\pm0.045$, compared with $0.504\pm0.045$ for plain MLP. Its AUROC and AUPRC are also higher: $0.806$ versus $0.768$ and $0.323$ versus $0.298$, respectively. The AUROC improvement is significant, while the differences in F1 and AUPRC are not. A possible reason is that ForEnt's test dataset is much smaller than UAV's, with 164 versus 7,726 windows, leading to less statistical power.} 

For the sequential alternatives, we train Bi-LSTM, Bi-GRU, and Bi-RNN uncertainty predictors on raw telemetry. On UAV, they achieve downstream F1 of $0.707\pm0.066$, $0.697\pm0.051$, and $0.695\pm0.068$, respectively, compared with $0.801\pm0.039$ for GatedMLP. All three F1 improvements are statistically significant ($p<0.001$), with large effect sizes ($\hat{A}_{12}=0.879$--$0.950$) \rev{On ForEnt, Bi-LSTM, Bi-GRU, and Bi-RNN achieve F1 of $0.467\pm0.070$, $0.503\pm0.068$, and $0.465\pm0.062$, respectively, compared with $0.522\pm0.045$ for GatedMLP. The F1 improvements over Bi-LSTM and Bi-RNN are statistically significant ($p\leq0.001$), with large effects sizes ($\hat{A}_{12}=0.760$--$0.764$). However, the improvement over Bi-GRU is not significant ($p=0.111$, $\hat{A}_{12}=0.617$).} These alternatives change both the input representation and the predictor architecture.

\subsubsection{RQ2.2: Ablation of \ulnr}
\label{subsubsec:ulnr}
We ablate \ulnr along two dimensions: (1) the sensitivity of the flip threshold $\tau$, and (2) a comparison against 14 established rebalancing strategies.

\noindent\textbf{Flip threshold $\tau$.} The threshold determines how unusual a safe window's uncertainty must be before it becomes eligible for relabeling. We vary $\tau$ from 0.5 to 4.0 to examine how the number of changed labels affects the precision--recall trade-off (Table~\ref{tab:flip_threshold}). On UAV, the number of relabeled windows decreases from 7,839 (11.30\% of the training set) at $\tau=0.5$ to 13 (0.02\%) at $\tau=4.0$. The main setting $\tau=3.0$ relabels 1,450 windows (2.09\%). \rev{On ForEnt, the selected $\tau=0.5$ relabels 129 windows (8.01\%). This number decreases to 50 (3.11\%) at $\tau=2.5$. No windows are relabeled at $\tau\geq3.0$ because the uncertainty z-scores remain below 3. F1 then falls to the plain predictor's $0.450\pm0.076$.}

\begin{table*}[t]
\centering
\color{black}
\captionsetup{textfont={color=black},labelfont={bf,color=black}}
\caption{Sensitivity to the flip threshold $\tau$ (RQ2.2). Values are mean $\pm$ standard deviation. Labels flipped counts safe training windows relabeled unsafe. Flip ratio is their proportion of training windows. Final ratio is the resulting unsafe proportion. Bold $\tau$ values identify the settings used in the main experiments: $3.0$ for UAV and $0.5$ for ForEnt. Bold numbers indicate the highest mean per dataset, including ties.}
\label{tab:flip_threshold}
\resizebox{0.97\textwidth}{!}{
\begin{tabular}{l c ccc ccccc}
\toprule
\textbf{Dataset} & $\bm{\tau}$ & \textbf{Labels flipped} & \textbf{Flip ratio} & \textbf{Final ratio} & \textbf{Precision} & \textbf{Recall} & \textbf{F1} & \textbf{AUROC} & \textbf{AUPRC} \\
\midrule
\multirow{8}{*}{\textbf{UAV}} & 0.5 & 7839$\pm$575 & 11.30$\pm$0.83\% & 12.62$\pm$0.83\% & 0.706$\pm$0.074 & 0.758$\pm$0.054 & 0.727$\pm$0.039 & 0.975$\pm$0.005 & 0.787$\pm$0.056 \\
 & 1.0 & 6867$\pm$466 & 9.90$\pm$0.67\% & 11.22$\pm$0.67\% & 0.712$\pm$0.078 & 0.760$\pm$0.059 & 0.730$\pm$0.034 & 0.976$\pm$0.005 & 0.794$\pm$0.047 \\
 & 1.5 & 5771$\pm$319 & 8.32$\pm$0.46\% & 9.64$\pm$0.46\% & 0.741$\pm$0.060 & 0.770$\pm$0.044 & 0.752$\pm$0.029 & 0.977$\pm$0.004 & 0.815$\pm$0.030 \\
 & 2.0 & 4409$\pm$144 & 6.36$\pm$0.21\% & 7.68$\pm$0.21\% & 0.743$\pm$0.078 & 0.759$\pm$0.046 & 0.748$\pm$0.047 & 0.976$\pm$0.006 & 0.813$\pm$0.047 \\
 & 2.5 & 2907$\pm$188 & 4.19$\pm$0.27\% & 5.51$\pm$0.27\% & 0.752$\pm$0.063 & 0.763$\pm$0.055 & 0.755$\pm$0.040 & \textbf{0.978$\pm$0.006} & \textbf{0.823$\pm$0.036} \\
 & \textbf{3.0} & 1450$\pm$424 & 2.09$\pm$0.61\% & 3.41$\pm$0.61\% & 0.792$\pm$0.064 & \textbf{0.814$\pm$0.051} & \textbf{0.801$\pm$0.039} & 0.972$\pm$0.005 & 0.776$\pm$0.033 \\
 & 3.5 & 372$\pm$333 & 0.54$\pm$0.48\% & 1.86$\pm$0.48\% & 0.779$\pm$0.083 & 0.616$\pm$0.103 & 0.678$\pm$0.054 & 0.958$\pm$0.020 & 0.747$\pm$0.067 \\
 & 4.0 & 13$\pm$35 & 0.02$\pm$0.05\% & 1.34$\pm$0.05\% & \textbf{0.819$\pm$0.088} & 0.526$\pm$0.072 & 0.633$\pm$0.035 & 0.939$\pm$0.014 & 0.682$\pm$0.041 \\
\midrule
\multirow{8}{*}{\textbf{ForEnt}} & \textbf{0.5} & 129$\pm$4 & 8.01$\pm$0.22\% & 31.80$\pm$0.22\% & 0.408$\pm$0.071 & \textbf{0.776$\pm$0.152} & 0.522$\pm$0.045 & \textbf{0.806$\pm$0.050} & \textbf{0.323$\pm$0.045} \\
 & 1.0 & 123$\pm$3 & 7.65$\pm$0.20\% & 31.43$\pm$0.20\% & 0.403$\pm$0.045 & 0.764$\pm$0.134 & 0.520$\pm$0.040 & 0.801$\pm$0.052 & 0.316$\pm$0.041 \\
 & 1.5 & 113$\pm$4 & 7.01$\pm$0.22\% & 30.79$\pm$0.22\% & \textbf{0.421$\pm$0.059} & 0.717$\pm$0.083 & \textbf{0.526$\pm$0.044} & 0.804$\pm$0.051 & 0.316$\pm$0.041 \\
 & 2.0 & 90$\pm$5 & 5.64$\pm$0.30\% & 29.42$\pm$0.30\% & 0.404$\pm$0.058 & 0.745$\pm$0.108 & 0.518$\pm$0.045 & 0.791$\pm$0.050 & 0.306$\pm$0.049 \\
 & 2.5 & 50$\pm$6 & 3.11$\pm$0.40\% & 26.90$\pm$0.40\% & 0.399$\pm$0.062 & 0.759$\pm$0.107 & 0.515$\pm$0.043 & 0.775$\pm$0.046 & 0.305$\pm$0.057 \\
 & 3.0 & 0 & 0.00\% & 23.79\% & 0.370$\pm$0.148 & 0.674$\pm$0.151 & 0.450$\pm$0.076 & 0.686$\pm$0.069 & 0.279$\pm$0.112 \\
 & 3.5 & 0 & 0.00\% & 23.79\% & 0.370$\pm$0.148 & 0.674$\pm$0.151 & 0.450$\pm$0.076 & 0.686$\pm$0.069 & 0.279$\pm$0.112 \\
 & 4.0 & 0 & 0.00\% & 23.79\% & 0.370$\pm$0.148 & 0.674$\pm$0.151 & 0.450$\pm$0.076 & 0.686$\pm$0.069 & 0.279$\pm$0.112 \\
\bottomrule
\end{tabular}
}
\end{table*}

\begin{table*}[t]
\centering
\color{black}
\captionsetup{textfont={color=black},labelfont={bf,color=black}}
\caption{Comparison with 14 alternative rebalancing methods (RQ2.2). Values are mean $\pm$ standard deviation. The statistical reference is \ulnr. Bold numbers indicate the highest mean per dataset, including ties.}
\label{tab:rq3}
\resizebox{0.84\textwidth}{!}{
\begin{tabular}{l l ccccc}
\toprule
\textbf{Dataset} & \textbf{Strategy} & \textbf{Precision} & \textbf{Recall} & \textbf{F1} & \textbf{AUROC} & \textbf{AUPRC} \\
\midrule
\multirow{15}{*}{\textbf{UAV}} & \ulnr & 0.792$\pm$0.064 & \textbf{0.814$\pm$0.051} & \textbf{0.801$\pm$0.039} & \textbf{0.972$\pm$0.005} & \textbf{0.776$\pm$0.033} \\
 & T-SMOTE & 0.715$\pm$0.124 & 0.524$\pm$0.068 & 0.599$\pm$0.074\textsuperscript{*}\,(L) & 0.904$\pm$0.029\textsuperscript{*}\,(L) & 0.625$\pm$0.089\textsuperscript{*}\,(L) \\
 & OSS & \textbf{0.852$\pm$0.050} & 0.503$\pm$0.022 & 0.631$\pm$0.020\textsuperscript{*}\,(L) & 0.931$\pm$0.015\textsuperscript{*}\,(L) & 0.694$\pm$0.029\textsuperscript{*}\,(L) \\
 & MiSLAS & 0.842$\pm$0.070 & 0.497$\pm$0.036 & 0.624$\pm$0.036\textsuperscript{*}\,(L) & 0.925$\pm$0.023\textsuperscript{*}\,(L) & 0.651$\pm$0.041\textsuperscript{*}\,(L) \\
 & ADASYN & 0.623$\pm$0.189 & 0.613$\pm$0.052 & 0.596$\pm$0.099\textsuperscript{*}\,(L) & 0.917$\pm$0.025\textsuperscript{*}\,(L) & 0.573$\pm$0.121\textsuperscript{*}\,(L) \\
 & GCL & 0.757$\pm$0.112 & 0.490$\pm$0.056 & 0.590$\pm$0.054\textsuperscript{*}\,(L) & 0.922$\pm$0.026\textsuperscript{*}\,(L) & 0.639$\pm$0.042\textsuperscript{*}\,(L) \\
 & RSG & 0.605$\pm$0.168 & 0.568$\pm$0.061 & 0.571$\pm$0.092\textsuperscript{*}\,(L) & 0.917$\pm$0.027\textsuperscript{*}\,(L) & 0.586$\pm$0.092\textsuperscript{*}\,(L) \\
 & SMOTE & 0.559$\pm$0.170 & 0.581$\pm$0.090 & 0.556$\pm$0.102\textsuperscript{*}\,(L) & 0.913$\pm$0.030\textsuperscript{*}\,(L) & 0.606$\pm$0.093\textsuperscript{*}\,(L) \\
 & Borderline-SMOTE & 0.524$\pm$0.163 & 0.594$\pm$0.051 & 0.540$\pm$0.094\textsuperscript{*}\,(L) & 0.916$\pm$0.028\textsuperscript{*}\,(L) & 0.540$\pm$0.116\textsuperscript{*}\,(L) \\
 & CC & 0.457$\pm$0.083 & 0.632$\pm$0.085 & 0.520$\pm$0.041\textsuperscript{*}\,(L) & 0.929$\pm$0.012\textsuperscript{*}\,(L) & 0.583$\pm$0.066\textsuperscript{*}\,(L) \\
 & RUS & 0.438$\pm$0.077 & 0.592$\pm$0.065 & 0.499$\pm$0.064\textsuperscript{*}\,(L) & 0.917$\pm$0.029\textsuperscript{*}\,(L) & 0.617$\pm$0.070\textsuperscript{*}\,(L) \\
 & SelMix & 0.637$\pm$0.093 & 0.415$\pm$0.030 & 0.499$\pm$0.036\textsuperscript{*}\,(L) & 0.814$\pm$0.025\textsuperscript{*}\,(L) & 0.490$\pm$0.043\textsuperscript{*}\,(L) \\
 & ReMix & 0.649$\pm$0.075 & 0.389$\pm$0.030 & 0.484$\pm$0.029\textsuperscript{*}\,(L) & 0.792$\pm$0.028\textsuperscript{*}\,(L) & 0.457$\pm$0.039\textsuperscript{*}\,(L) \\
 & LDAM-DRW & 0.822$\pm$0.085 & 0.507$\pm$0.046 & 0.623$\pm$0.032\textsuperscript{*}\,(L) & 0.927$\pm$0.025\textsuperscript{*}\,(L) & 0.641$\pm$0.063\textsuperscript{*}\,(L) \\
 & CW & 0.840$\pm$0.069 & 0.503$\pm$0.026 & 0.627$\pm$0.024\textsuperscript{*}\,(L) & 0.931$\pm$0.023\textsuperscript{*}\,(L) & 0.663$\pm$0.038\textsuperscript{*}\,(L) \\
\midrule
\multirow{15}{*}{\textbf{ForEnt}} & \ulnr & \textbf{0.408$\pm$0.071} & 0.776$\pm$0.152 & \textbf{0.522$\pm$0.045} & \textbf{0.806$\pm$0.050} & \textbf{0.323$\pm$0.045} \\
 & T-SMOTE & 0.300$\pm$0.125 & 0.626$\pm$0.167 & 0.379$\pm$0.098\textsuperscript{*}\,(L) & 0.617$\pm$0.103\textsuperscript{*}\,(L) & 0.226$\pm$0.099\textsuperscript{*}\,(L) \\
 & OSS & 0.353$\pm$0.088 & 0.680$\pm$0.116 & 0.452$\pm$0.069\textsuperscript{*}\,(L) & 0.694$\pm$0.082\textsuperscript{*}\,(L) & 0.277$\pm$0.111\textsuperscript{*}\,(M) \\
 & MiSLAS & 0.331$\pm$0.062 & 0.715$\pm$0.103 & 0.445$\pm$0.050\textsuperscript{*}\,(L) & 0.684$\pm$0.059\textsuperscript{*}\,(L) & 0.256$\pm$0.091\textsuperscript{*}\,(L) \\
 & ADASYN & 0.317$\pm$0.131 & 0.597$\pm$0.167 & 0.386$\pm$0.086\textsuperscript{*}\,(L) & 0.635$\pm$0.074\textsuperscript{*}\,(L) & 0.250$\pm$0.081\textsuperscript{*}\,(L) \\
 & GCL & 0.339$\pm$0.118 & 0.688$\pm$0.102 & 0.437$\pm$0.056\textsuperscript{*}\,(L) & 0.672$\pm$0.063\textsuperscript{*}\,(L) & 0.247$\pm$0.083\textsuperscript{*}\,(L) \\
 & RSG & 0.344$\pm$0.085 & 0.664$\pm$0.103 & 0.442$\pm$0.073\textsuperscript{*}\,(L) & 0.666$\pm$0.058\textsuperscript{*}\,(L) & 0.252$\pm$0.067\textsuperscript{*}\,(L) \\
 & SMOTE & 0.337$\pm$0.128 & 0.605$\pm$0.129 & 0.408$\pm$0.099\textsuperscript{*}\,(L) & 0.640$\pm$0.080\textsuperscript{*}\,(L) & 0.276$\pm$0.125\textsuperscript{*}\,(L) \\
 & Borderline-SMOTE & 0.307$\pm$0.126 & 0.620$\pm$0.134 & 0.385$\pm$0.090\textsuperscript{*}\,(L) & 0.616$\pm$0.066\textsuperscript{*}\,(L) & 0.229$\pm$0.086\textsuperscript{*}\,(L) \\
 & CC & 0.222$\pm$0.104 & \textbf{0.797$\pm$0.269} & 0.305$\pm$0.079\textsuperscript{*}\,(L) & 0.418$\pm$0.143\textsuperscript{*}\,(L) & 0.155$\pm$0.062\textsuperscript{*}\,(L) \\
 & RUS & 0.381$\pm$0.071 & 0.698$\pm$0.095 & 0.485$\pm$0.055\textsuperscript{*}\,(M) & 0.725$\pm$0.051\textsuperscript{*}\,(L) & 0.290$\pm$0.068\textsuperscript{*}\,(L) \\
 & SelMix & 0.361$\pm$0.064 & 0.721$\pm$0.116 & 0.471$\pm$0.043\textsuperscript{*}\,(L) & 0.733$\pm$0.054\textsuperscript{*}\,(L) & 0.276$\pm$0.052\textsuperscript{*}\,(M) \\
 & ReMix & 0.358$\pm$0.074 & 0.744$\pm$0.105 & 0.474$\pm$0.060\textsuperscript{*}\,(L) & 0.731$\pm$0.064\textsuperscript{*}\,(L) & 0.268$\pm$0.063\textsuperscript{*}\,(L) \\
 & LDAM-DRW & 0.399$\pm$0.060 & 0.682$\pm$0.035 & 0.499$\pm$0.041\textsuperscript{*}\,(L) & 0.741$\pm$0.049\textsuperscript{*}\,(M) & 0.304$\pm$0.071\textsuperscript{*}\,(M) \\
 & CW & 0.350$\pm$0.136 & 0.679$\pm$0.142 & 0.437$\pm$0.054\textsuperscript{*}\,(L) & 0.681$\pm$0.050\textsuperscript{*}\,(L) & 0.266$\pm$0.082\textsuperscript{*}\,(L) \\
\bottomrule
\end{tabular}
}
\par\smallskip
\begin{minipage}{0.84\textwidth}\footnotesize \textsuperscript{*}$p<0.05$ for the reference method scoring higher (one-sided paired Wilcoxon). (N)/(S)/(M)/(L): negligible/small/medium/large $\hat{A}_{12}$ magnitude.
\end{minipage}
\end{table*}

\noindent\textbf{Alternative Rebalancing Approaches.} 
\rev{\ulnr achieves the highest F1, AUROC, and AUPRC among the 14 alternative rebalancing methods on both datasets (Table~\ref{tab:rq3}). On UAV, its F1 of $0.801\pm0.039$ exceeds OSS ($0.631\pm0.020$), the strongest alternative, by 17.0 percentage points. On ForEnt, its F1 of $0.522\pm0.045$ exceeds LDAM-DRW ($0.499\pm0.041$), the strongest alternative, by 2.3 percentage points. Its ForEnt AUROC and AUPRC are $0.806\pm0.050$ and $0.323\pm0.045$, compared with $0.741\pm0.049$ and $0.304\pm0.071$ for LDAM-DRW.}

\rev{On UAV, the F1, AUROC, and AUPRC improvements over all 14 alternatives are significant ($p<0.05$), with large effects.}

\rev{On ForEnt, the F1, AUROC, and AUPRC improvements over all 14 alternatives are significant ($p<0.05$). F1 effects are large except against RUS (medium). AUROC effects are large except against LDAM-DRW (medium). AUPRC effects are large except against OSS, SelMix, and LDAM-DRW (medium).}

\noindent\textbf{A Plausible Explanation for \ulnr's Performance.}
\rev{We hypothesize that \ulnr's improvements come from relabeling samples near the boundary between safe and unsafe operation. We therefore compare the distributions of measurements related to safety in flipped and unflipped safe training windows: minimum obstacle distance on UAV and mean absolute roll (body tilt) over each 5-s window on ForEnt.}

\rev{On UAV, flipped windows have a mean minimum obstacle distance of $1.90\pm0.04\,\mathrm{m}$, compared with $10.93\pm0.02\,\mathrm{m}$ for unflipped windows. The flipped group's mean is only $0.40\,\mathrm{m}$ above the $1.5\,\mathrm{m}$ safety threshold. Originally unsafe windows have a mean distance of $0.87\,\mathrm{m}$. On ForEnt, the mean absolute roll is $0.063\pm0.001\,\mathrm{rad}$ for flipped windows and $0.038\pm0.000\,\mathrm{rad}$ for unflipped windows. The tilt of flipped windows is close to the mean of $0.066\,\mathrm{rad}$ observed during entrapment. The flipped windows have significantly lower obstacle distances on UAV and greater body tilt on ForEnt than unflipped windows.}

\rev{These results provide evidence for our hypothesis that \ulnr selects samples near the safe/unsafe boundary. These windows also have high behavioral uncertainty, so their safe labels may be noisy. LNR~\cite{lnr} shows that suitable label flips can improve learning from imbalanced data. Relabeling these uncertain boundary samples may therefore reduce overfitting to noisy safety labels and improve safety prediction.}

\begin{tcolorbox}
\rev{GatedMLP achieves higher F1, AUROC, and AUPRC than both the direct statistical scores and alternative predictor architectures on both datasets. \ulnr also achieves the highest F1, AUROC, and AUPRC among the alternative rebalancing methods on both datasets.}
\end{tcolorbox}

\section{Threats to Validity}

\noindent\textbf{Internal Validity.} \rev{Differences in predictor architecture or training procedure could affect the improvements attributed to \ulnr. We control for these factors by comparing \method with the same Bi-LSTM trained on the original labels. The ablations of the uncertainty predictor also assess alternative uncertainty scores and architectures. We selected Bi-LSTM through preliminary experiments on held-out validation data. With this safety predictor, \method achieves the highest F1, AUROC, and AUPRC among the 14 safety-prediction baselines on both datasets. Our evaluation does not cover all possible architectures for the safety predictor. Other architectures may further improve performance, and we leave their evaluation for future work.}

Unequal hyperparameter tuning could favor some methods. On each dataset, we use held-out validation data and give each prediction model the same budget of candidate configurations. We provide the search spaces, implementation defaults, and selected configurations in our repository.

\noindent\textbf{Conclusion Validity.} \rev{Random initialization can affect the reliability of the comparisons. We mitigate this threat by repeating each experiment and reporting statistical tests and effect sizes, as described in Section~\ref{subsec:metrics}. ForEnt's test dataset contains 164 windows and 22 unsafe windows, which limits the precision of its performance estimates.}

\noindent\textbf{External Validity.} \rev{Our results may not generalize to other CPS platforms or operating environments. We mitigate this threat by evaluating \method on UAV collision monitoring and ForEnt ground-robot entrapment detection. The datasets differ in telemetry channels, unsafe outcomes, class balance, and availability of uncertainty labels. ForEnt provides no uncertainty labels, so we obtain them through automated annotation based on horizontal-acceleration energy. Its results show applicability when these labels are not already available. Evaluation on additional CPS domains and with other annotation rules remains future work.}

\section{Related Work}

In this section, we discuss related works on safety monitoring in CPS (Section~\ref{subsec:cps_safety_rw}) and dataset rebalancing techniques (Section~\ref{subsec: data_bal_rw}).

\subsection{Safety Monitoring in CPS}
\label{subsec:cps_safety_rw}

\noindent Safety monitoring in CPS detects hazardous system states during operation and supports intervention to avoid safety violations. Prior research includes model-based and data-driven approaches.

Model-based approaches express safety knowledge through control models or formal specifications. STPA~\cite{leveson2014comparison} analyzes hazards arising from unsafe control interactions. Signal Temporal Logic (STL) specifies properties over observed signals~\cite{maler2004monitoring}. Robustness measures quantify the margin of satisfaction or violation~\cite{donze2010robust}. These approaches require safety properties or control abstractions to be specified for the system under study.

Data-driven monitors learn patterns from operational data. Anomaly detectors identify deviations from nominal behavior~\cite{chandola2009anomaly,superialist}, although a deviation alone does not establish a safety violation. Other monitors use learned predictions together with safety constraints. \rev{Sharifi et al.~\cite{sharifi2025system} forecast safety metrics. Lin et al.~\cite{lin2025safety} combine trajectory prediction with safety monitoring under distribution shift. Catak et al.~\cite{catak2022uncertainty} use uncertainty to predict whether another model's output is incorrect.}

Our work focuses on supervised window-level safety prediction when unsafe training examples are scarce. It uses behavioral uncertainty to select training labels for rebalancing, then learns the runtime safety predictor from the rebalanced data.

\subsection{Dataset Rebalancing}
\label{subsec: data_bal_rw}

Existing approaches to imbalanced learning are commonly grouped into two categories: data-level and algorithm-level methods~\cite{johnson2019survey,chen2024survey}.

Algorithm-level methods address imbalance by modifying the training objective to mitigate bias without altering the training data. Cost-sensitive learning penalizes misclassification of minority-class samples more heavily during training~\cite{elkan2001foundations}. Class weighting scales the loss contribution of each class, for example by assigning larger loss weights to rarer classes~\cite{khan2017cost, zhu2018class}. Cui et al.~\cite{cui2019classbalanced} use the inverse effective number of samples as the class weight to account for overlap among observations.  LDAM-DRW~\cite{cao2019learning} assigns larger classification margins to classes with fewer training samples, increasing separation between minority and majority classes, and defers class re-weighting to a later stage of training, allowing the model to first learn an informative feature representation before rebalancing. Gaussian Clouded Logit adjustment (GCL)~\cite{li2022long} applies class-dependent Gaussian perturbations to classifier logits. Mixup and Label-Aware Smoothing (MiSLAS)~\cite{zhong2021improving} applies label-aware smoothing that varies smoothing intensity per class based on the number of samples in that class, reducing overconfident predictions for majority classes.

Data-level methods rebalance class distributions through \textit{oversampling} or \textit{undersampling}. \textit{Oversampling} increases minority-class samples. For instance, SMOTE \cite{chawla2002smote} generates synthetic minority samples by interpolating existing examples, while ADASYN~\cite{he2008adasyn} extends SMOTE by generating more synthetic samples for minority class examples that are challenging to classify. Borderline-SMOTE~\cite{han2005borderline} generates samples near the class boundary, while Rare-class Sample Generator (RSG)~\cite{wang2021rsg} synthesizes minority-class samples in feature space based on patterns learned from majority-class data. T-SMOTE~\cite{zhao2022t}, an oversampling method for imbalanced time-series classification, adapts SMOTE to time-series data by generating synthetic minority samples that preserve temporal structure, particularly near class boundaries. Mixup-based methods such as ReMix~\cite{chou2020remix} and SelMix~\cite{ramasubramanian2024selective} adapt interpolation between samples~\cite{zhang2017mixup} to imbalanced learning and objective-aware fine-tuning, respectively. In contrast, \textit{undersampling} reduces the number of majority-class samples. Undersampling methods include One-Sided Selection (OSS)~\cite{batista2000applying}, which combines nearest-neighbor subset selection with Tomek-link cleaning, and Cluster Centroids (CC)~\cite{macqueen1967multivariate}, which reduces the majority class by replacing each cluster with its centroid.

LNR~\cite{lnr} uses classifier scores to guide stochastic majority-to-minority label changes, addressing imbalance without generating or removing samples. Its evaluation includes tabular and image classification. Our work applies probabilistic label rebalancing to CPS telemetry windows, using behavioral uncertainty as the signal for selecting safe windows to relabel.

\section{Conclusion and Future Work}
\label{sec:conclusion}

In this paper, we propose \method, a safety monitoring approach for Cyber-Physical Systems (CPSs). \method first trains an uncertainty predictor that summarizes each telemetry window into distributional kinematic features and outputs an uncertainty score. It then applies an uncertainty-guided LNR (\ulnr) mechanism that probabilistically relabels high-uncertainty \{safe\} training windows as \{unsafe\}, increasing minority-class representation without synthesizing new data. Finally, it trains a safety predictor on the rebalanced dataset for runtime monitoring.

\rev{Across the UAV and ForEnt case studies, \method achieves F1/AUROC/AUPRC of $0.801/0.972/0.776$ and $0.522/0.806/0.323$, respectively. Relative to the strongest safety-prediction baselines, it improves F1/AUROC/AUPRC by 14.3/2.2/6.1 percentage points on UAV and 4.4/9.2/1.4 on ForEnt. Relative to the strongest alternative rebalancing methods, the corresponding gains are 17.0/4.1/8.2 and 2.3/6.5/1.9 percentage points. Uncertainty-guided relabeling also outperforms Early Fusion and Late Fusion on all three metrics. GatedMLP yields the best uncertainty scores among the tested alternatives on both datasets, improving F1, AUROC, and AUPRC.}

\rev{These improvements indicate that \method is effective under both severe and moderate class imbalance. The F1 gain is larger on UAV than on ForEnt (14.3 versus 4.4 percentage points), suggesting that, as expected, rebalancing is more beneficial when unsafe training samples are scarcer.}

On UAV, relabeled \{safe\} windows exhibit lower obstacle clearance than unchanged \{safe\} windows; on ForEnt, they exhibit greater body tilt. Selecting these physically riskier windows plausibly explains the observed improvements.

Future work includes evaluating \method in additional CPS domains (e.g., autonomous submarines) to assess generalizability across sensor modalities and operating conditions. We also plan to study predictive uncertainty estimation techniques such as MC Dropout and Deep Ensembles to understand how uncertainty quality affects rebalancing effectiveness.

\section{Acknowledgment}

This publication has emanated from research conducted with the financial support of Taighde Éireann – Research Ireland under Grant number 13/RC/2094\_2. Lionel Briand was also supported by the NSERC Discovery Grant program. For the purpose of Open Access, the authors have applied a CC BY public copyright license to any Author Accepted Manuscript version arising from this submission.

\bibliographystyle{IEEEtran}
\bibliography{references}

\end{document}